\documentclass[manuscript, screen]{acmart}
\usepackage{booktabs} 

\usepackage[ruled]{algorithm2e} 

\usepackage{caption}
\usepackage{graphicx}
\usepackage{subfigure}
\usepackage{float}
\usepackage{enumerate}
\usepackage{empheq}
\usepackage{amsmath}
\usepackage{amssymb}
\usepackage{amsfonts}
\usepackage{textcomp}
\usepackage{tabularx}
\usepackage{booktabs}
\usepackage{mathtools}
\usepackage{multirow}
\usepackage{soul}
\usepackage{array}
\usepackage{color}
\usepackage{caption}
\usepackage{dsfont}
\usepackage[flushleft]{threeparttable}

\setcopyright{acmcopyright}

\acmDOI{0000001.0000001}

\begin{document}
\title{Physics-Guided Machine Learning for Scientific Discovery: An Application in Simulating Lake Temperature Profiles}

\author{Xiaowei Jia}
\affiliation{%
  \institution{University of Minnesota, Twin Cities}
\department{Department of Computer Science and Engineering}
\city{Minneapolis}
\state{MN}
\postcode{55455}
\country{USA}
}
\email{jiaxx221@umn.edu}
\author{Jared Willard}
\affiliation{%
  \institution{University of Minnesota, Twin Cities}
 \department{Department of Computer Science and Engineering}
\city{Minneapolis}
\state{MN}
\postcode{55455}
\country{USA}
}
\email{willa099@umn.edu}
\author{Anuj Karpatne}
\affiliation{%
 \institution{Virginia Tech}
  \department{Department of Computer Science}
  \city{Blacksburg}
  \state{VA}
  \postcode{24061}
  \country{USA}
}
\email{karpatne@vt.edu}
\author{Jordan S Read}
\affiliation{%
  \institution{U.S. Geological Survey}
\country{USA}
}
\email{jread@usgs.gov}
\author{Jacob A Zwart}
\affiliation{%
  \institution{U.S. Geological Survey}
\country{USA}
}
\email{jzwart@usgs.gov}
\author{Michael Steinbach}
\affiliation{%
  \institution{University of Minnesota}
\department{Department of Computer Science and Engineering}
\city{Minneapolis}
\state{MN}
\postcode{55455}
\country{USA}
}
\email{stei0062@umn.edu}
\author{Vipin Kumar}
\affiliation{%
  \institution{University of Minnesota, Twin Cities}
\department{Department of Computer Science and Engineering}
\city{Minneapolis}
\state{MN}
\postcode{55455}
\country{USA}
}
\email{kumar001@umn.edu}

\renewcommand\shortauthors{Jia, X. et al}

\begin{abstract}
{P}hysics-based models  are often used to study engineering and environmental systems. {The ability to model these  systems is the key to achieving our future environmental sustainability and improving the
quality of human life. This paper focuses on simulating lake water temperature, which is critical for understanding the impact of changing climate on aquatic ecosystems  and assisting in aquatic resource management decisions. General Lake Model (GLM) is a state-of-the-art physics-based model used for addressing such problems.} 
{However, like other physics-based models used for studying scientific and engineering systems, it has} several well-known limitations due to simplified representations of the physical processes being modeled or challenges in selecting appropriate parameters.   
While-state-of-the-art machine learning models can sometimes 
outperform physics-based models given ample amount of training data, they can produce results that are physically inconsistent. 
This paper proposes a physics-guided recurrent neural network model (PGRNN) that combines RNNs and physics-based models to leverage their complementary strengths and improves the modeling of physical processes. Specifically, we show that a PGRNN can improve prediction accuracy over that of physics-based models {(by over 20\% even with very few training data)}, while generating outputs consistent with physical laws. 
An important aspect of our PGRNN approach lies in its ability to incorporate the knowledge encoded in physics-based models. This allows training the PGRNN model using very few true observed data while also ensuring high prediction accuracy.    
Although we present and evaluate this methodology in the context of modeling the dynamics of temperature in lakes, it is applicable more widely to a range of scientific and engineering disciplines where physics-based (also known as mechanistic) models are used.
\end{abstract}

%
%


%
%


\maketitle


\section{Introduction}

Physics-based models have been widely used to study engineering and environmental systems in domains such as hydrology, climate science, materials science, agriculture, and computational chemistry. Despite their extensive use, these models have several well-known limitations due to simplified representations of the physical processes being modeled or challenges in selecting appropriate parameters. 
There is a tremendous opportunity to systematically advance modeling in these domains by using machine learning (ML) methods. However, capturing this opportunity is contingent on a paradigm shift in data-intensive scientific discovery since the "black box" use of ML often leads to serious false discoveries in scientific applications \cite{lazer2014parable,karpatne2017theory}. In this paper, we present a novel methodology for combining physics-based models with state-of-the-art deep learning methods to leverage their complementary strengths. 

Even though physics-based models are based on known physical laws that govern relationships between input and output variables, the majority of physics-based models are necessarily approximations of reality due to incomplete knowledge of certain processes,  which introduces bias. In addition, they often contain a large number of parameters whose values must be estimated with the help of limited observed data. A standard approach for calibrating these parameters is to 
exhaustively search the space of parameter combinations and choose parameter combinations that result in the best performance on training data. Besides its computational cost, this approach is 
also prone to over-fitting due to heterogeneity in the underlying processes in both space and time. 
The limitations of physics-based models cut across discipline boundaries and are well known in the scientific community; e.g., see a series of debate papers in hydrology~\cite{lall2014debates, gupta2014debates, mcdonnell2014debates}. 

ML models, given their tremendous success in several commercial applications (e.g., computer vision, and natural language processing) are increasingly being considered as promising alternatives to physics-based models by the scientific community. State of the art (SOA) ML models {(e.g., Long-Short Term Memory (LSTM), Convolutional Neural Networks (CNN)), and the attention mechanism)}  given enough data, 
can often perform better than traditional empirical models (e.g., regression-based models) used by science communities as an alternative to physics-based models~\cite{graham2008big, goh2017deep}. However, direct application of black-box ML models to a scientific problem encounters {several} major challenges: 
{1. Effective modeling of physical processes (that may be unfolding and interacting at multiple scales in space and time) is dependent on the capacity of ML models in extracting complex patterns from data.}
2. {Training ML models} requires a lot of {labeled}  data, which is scarce in most practical settings {given the substantial human efforts and material cost required to deploy and maintain sensors. } 
3. Empirical models (including the SOA ML models) simply identify statistical relations between inputs and the system variables of interest (e.g., the temperature profile of the lake) without taking into account any physical laws (e.g., conservation of energy or mass) and thus can produce results that are inconsistent with physical laws. {Hence, even if they produce accurate predictions, such models cannot be used in practice by domain experts and other stakeholders. }
4. Relationships produced by empirical models can at best be valid only for the set of variable combinations present in the training data and are unable to generalize to scenarios unseen in the training data. 
For example, a ML model trained for today's climate may not be accurate for future warmer climate scenarios.

The goal of this work is to improve the modeling of engineering and environmental systems. 
Effective representation of physical processes in such systems will require development of novel abstractions and architectures. 
In addition, the optimization process to produce an ML model will have to consider not just accuracy (i.e., how well the output matches the observations) but also its ability to provide physically consistent results. 
{The most common approach for directly addressing the imperfection of physics-based models in the scientific community is residual modeling, where an ML model learns to predict the errors, or residuals, made by a physics-based model~\cite{kani2017dr,wan2018data,san2018machine}. One of the key limitations of these approaches is their inability to provide predictions that are consistent with known physical laws. Karpatne et al.~\cite{karpatne2017physics} further extends residual modeling by using simulated data as additional input to the ML model. 3. This new framework  permits incorporation  of
physical constraints that can be defined purely on the output of the model ~\cite{karpatne2017physics,muralidhar2018incorporating,beucler2019achieving}. However, these methods cannot incorporate more general constraints that are based on internal states of the physical system (e.g., energy conservation).  In addition, all of these approaches still require a lot of   training data, and thus cannot address the data scarcity challenge.}

{In this paper}, we present Physics-Guided Recurrent Neural Network models (PGRNN) as a general framework for modeling physical phenomena with potential applications for many disciplines. 
The PGRNN model has a number of novel aspects: 

1. Many temporal processes in environmental/engineering systems involve complex long-term temporal dependencies  that cannot be captured by a plain neural network or a simple temporal model such as a standard RNN. In contrast, in PGRNN we use advanced ML models such as LSTM, {which use the internal memory structure to preserve long-term temporal dependencies and thus has the potential to capture complex physical patterns that last over several months or years.} 

2. The proposed PGRNN can incorporate explicit physical laws such as energy conservation or mass conservation.   
This is done by introducing additional variables in the recurrent structure to keep track of physical states that can be used to check for consistency with physical laws. 
In addition, we generalize the loss function to include a physics-based penalty~\cite{karpatne2017theory}.  Thus, the overall training loss is 
$\mathcal{L} = \text{Supervised loss} (Y_{pred},Y_{true})+\text{Physics-based Penalty}$, where the first term on the right hand side represents the supervised training loss between the predicted outputs $Y_{pred}$ and the observed outputs $Y_{true}$ (e.g., RMSE in regression or cross-entropy in classification), 
and the {second} term represents the physical consistency-based penalty. In addition, to favoring physically consistent solutions, another major side benefit of including physics-based  penalty in the loss function is that it can be applied even to instances for which output (observed) data is not available since the physics-based penalty can be computed as long as input (driver) data is available. Note that in absence of physics based penalty, training loss can be computed only on those time steps where observed output is available.  Inclusion of physics based loss term allows a much more robust training, especially in situations, where observed output is available on only a small number of time steps.  

3. 
Physics based/mechanistic models contain a  lot of domain knowledge that goes well beyond what can be captured as constraints such conservation laws.  To leverage this knowledge, we generate a large amount of ``synthetic''  observation data by executing physics based models for a variety input drivers (that are easily available) and use the synthetic observation to pre-train the ML model. The idea here is that training from synthetic data generated by imperfect physical models may allow the ML model to get close enough to the target solution, so only a small amount of observed data (ground truth labels) is needed to further refine the model. In addition, the synthetic data is guaranteed to be physically consistent due to the nature of the process model being founded on physical principles.  

Our proposed Physics-Guided Recurrent Neural Networks model (PGRNN) is developed for the purpose of predicting lake water temperatures at various depths at the daily scale. The temperature of water in a lake is known to be an ecological ``master factor"~\cite{magnuson1979temperature} that controls the growth, survival, and reproduction of fish~\cite{roberts2013fragmentation}. Warming water temperatures can increase the occurrence of aquatic invasive species~\cite{rahel2008assessing,roberts2017nonnative}, which may displace fish and native aquatic organisms, result in more harmful algal blooms (HABs)~\cite{harris2017predicting,paerl2008blooms}. Understanding temperature change and the resulting biotic ``winners and losers" is timely science that can also be directly applied to inform priority action for natural resources. 
Given the importance of this problem, the aquatic science community has developed numerous {models} for the simulation of temperature, including the General Lake Model (GLM)~\cite{hipsey2019general}, which simulates the physical processes (e.g., vertical mixing, and the warming or cooling of water via energy lost or gained from fluxes such as solar radiation and evaporation, etc.). As is typical for any such model, GLM is only an approximation of the physical reality, and has a number of parameters (e.g., water clarity, mixing efficiency, and wind sheltering) that often need to be calibrated using observations.

We evaluate the proposed PGRNN method in a real-world system, Lake Mendota (Wisconsin), 
which is one of the most extensively studied lake systems in the world. We chose this lake because it has 
plenty of observed data that can be used to evaluate the performance of any new approach.   
In particular, we can measure the performance of different algorithms by varying the the amount of observations used for training. This helps test the effectiveness of the proposed methods in data-scarce scenarios, which is important since most real-world lakes 
have very few observations or are not observed at all (they usually have less than 1\% of observations that are available for Mendota). 
In addition, Lake Mendota is large and deep enough such that it shows a variety of temperature patterns (e.g., stratified temperature patterns in warmer seasons and well-mixed patterns in colder seasons). This allows us to test the capacity of ML models in capturing such complex temperature patterns.

Our main contributions are as follows. We show that it is possible to effectively model the temporal dynamics of temperature in lakes using LSTMs provided that enough observed data is available for training. We show that traditional LSTMs can be augmented to take energy conservation into account and track the balance of energy loss and gain relative to temperature change (a physical law of thermodynamics). Including such components in models to make the output consistent with physical laws can make them more acceptable for use by scientists and also may 
improve the prediction performance. 
We also studied the benefit of pre-training this model using synthetic data (i.e., the output of a generic physics-based model)  
and then refining it using only a small amount of observation data. The results show that 
such pre-trained models can easily outperform the state-of-the art physics-based model by using a small amount of observed data. Moreover, we show that such pre-training is useful even if it uses  
simulated data from lakes that are very different in geometry, clarity or climate 
than the lake being studied.  
These results confirm that the PGRNN can leverage the strengths of physics-based models while also filling in knowledge gaps  
by overlaying features learned from data. 

{The proposed method  has general applicability to many
scientific applications. In fact its effectiveness has already been shown in two different applications in aquatic science~\cite{read2019process,hanson2020predicting}.  As discussed in~\cite{willard2020integrating}, the overall approach is applicable to a wide range of domains such as hydrology, Computational fluid dynamics (CFD), and crop modeling.}


The organization of the paper is as follows: In Section 2, we describe the preliminary knowledge and the setting of our problem. Section 3 presents the discussions related to the proposed PGRNN model. In Section 4, we extensively evaluate the proposed method in a real-world dataset. We then recapitulate related existing work in Section 5 before we conclude our work in Section 6. A preliminary version of this work appeared in~\cite{jia2019physics}.

\section{Problem formulation and preliminaries}

\subsection{Problem formulation}
Our goal is to simulate the temperature of water in the lake at each depth $d$, and on each date $t$, given physical variables governing the dynamics of lake temperature. This problem is referred to
as 1D-modeling of temperature (depth being the single
dimension). 
Specifically, $x_t$ represents input physical variables at on a specific date $t$,  
which include meteorological recordings at the surface of water 
such as the amount of solar radiation (in W/m$^2$, for short-wave and long-wave), wind speed (in m/s), air temperature (in $^\circ C$), relative humidity (0-100\%), rain (in cm), snow indicator (True or False), as well as the value of depth (in m) and day of year (1-366).  
These chosen features are known to be the primary drivers of lake thermodynamics \cite{hipsey2019general}. 
Given these input drivers $x_t$ and a depth level $d$, we aim to predict water temperature $\{y_{d,t}\}_{t=1}^T$ at this depth over the entire study period. For simplicity, we use $x_t$ and $y_t$ to represent $\{x_t,d\}$ and $y_{d,t}$ in the paper when it causes no ambiguity.  
During the training process, we are given the sparse ground-truth observed temperature profiles on certain dates and at certain depths captured by in-water sensors (more dataset description is provided in Section~\ref{sec:dataset}). 


\begin{figure} [!h]
\centering
\includegraphics[width=0.85\columnwidth]{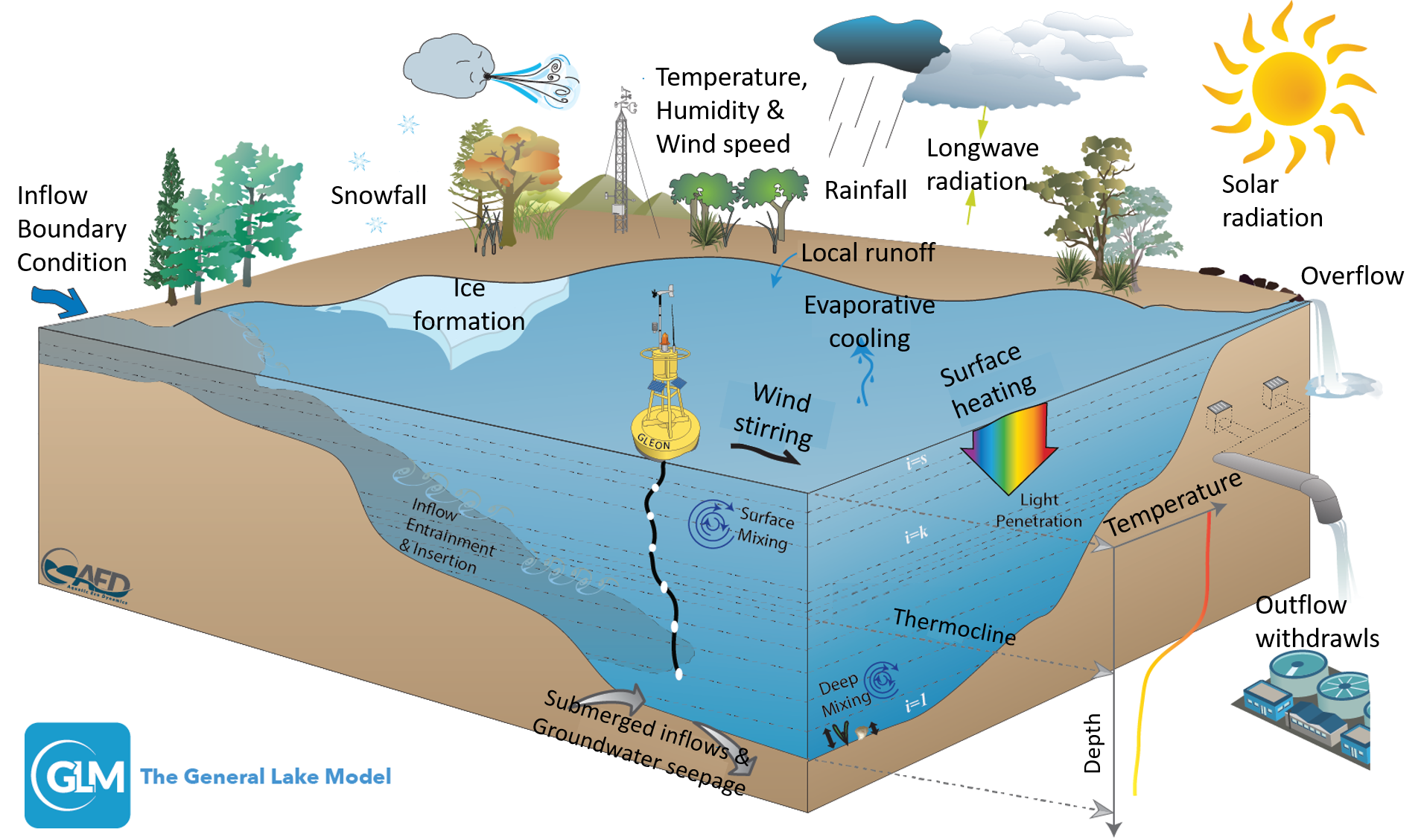}
\caption{A pictorial description of the physical processes simulated by the General Lake Model~\cite{hipsey2014glm}. These processes govern the dynamics of temperature in a lake.}
\label{GLM}
\end{figure}
\subsection{General Lake Model (GLM)}

The physics-based GLM captures a variety of physical processes governing
the dynamics of water temperature in a lake, including the heating
of the water surface due to incoming short-wave radiation, the attenuation of radiation beneath the water surface, the mixing of layers with varying thermal energy
at different depths, and the loss of heat from
the surface of the lake via evaporation or outgoing long-wave
radiation (shown in Fig.~\ref{GLM}). We use GLM
as our preferred physics-based model for lake
temperature modeling due to its model performance and wide use among the lake modeling community.

The GLM has a number of parameters (e.g., parameters related to vertical mixing, wind sheltering, and water clarity) that are often calibrated specifically to individual lakes if training data are available. The basic calibration method (common to a wide range of scientific and engineering problems) is to run the model for combinations of parameter values and select the parameter set that minimizes model error. This calibration process can be both labor- and computationally-intensive. 
Furthermore, the calibration process, applied even in the presence of ample training data, 
is still limited by simplifications and rigid formulations in these physics-based models.

\subsection{Machine learning model for sequential data}

{There is a class of ML models that}  aims to learn a black-box transformation from the input series $\{x_1,x_2,...,x_T\}$ to target variables $\{y_1,y_2,...,y_T\}$. In this problem, {the observation data can be sparse for certain depths so it is infeasible to train individual models for each depth separately. Instead,} we will train a {global} model that {uses depth as an input feature.
} This {makes it possible to use observation data from any depth at any time step} {for training this model}. Later in Section 4 we will show that {such a global model}  can still very well capture temporal dynamics at each depth separately.

We also use area-depth profile as additional information to compute energy constraints (see Section~\ref{sec:energy-method}). Since we train machine learning models that are specific to a target lake, the area-depth profile remains the same on different days and thus we do not include it in the input features.

\section{Our Proposed PGRNN method}
In this section, we will discuss the proposed PGRNN model in detail. First, we describe how to train an LSTM to model temperature dynamics using sparse observed data.  Second, we describe how to combine the energy conservation law and the standard recurrent neural networks model. Then, we further utilize a pre-training method to improve the learning performance even with limited training data. 

\subsection{Recurrent Neural Networks and Long-Short Term Memory Networks}
%

Recent advances in deep learning models enable automatic extraction of representative patterns from multivariate input temporal data to better predict the target variable. As one of the most popular temporal deep learning models, RNN models have shown success in a broad range of applications. The power of the RNN model lies in its ability to combine the input data at the current and previous time steps to extract an  informative hidden representation  $h_t$. In an RNN, the hidden representation $h_t$ is generated using the following equation:
\begin{equation}
h_t = \text{tanh}(W_h h_{t-1}+W_x x_{t}),
\label{h}
\end{equation}
where $W_{h}$ and $W_x$ represent the weight matrices that connect $h_{t-1}$ and $x_t$, respectively. Here the bias terms are omitted as they can be absorbed into the weight matrix.

While RNN models can model transitions across time, they gradually lose the connections to long histories as time progresses~\cite{bengio1994learning}. Therefore, the RNN-based method may fail to grasp long-term patterns that are common in scientific applications. For example, the seasonal patterns and yearly patterns that commonly exist in environmental systems can last for many time steps if we use data at a daily scale. The standard RNN fails to memorize long-term temporal patterns because it does not explicitly generate a long-term memory to store previous information but only captures the transition patterns between consecutive time steps. It is well-known~\cite{chen1992dynamic, pan2018longterm} that such issue of memory is a major difficulty in the study of dynamical system.   

As an extended version of the RNN, LSTM is better in modeling long-term dependencies where each time step needs more contextual information from the past. 
The difference between LSTM and RNN lies in the generation of the hidden representation $h_t$. In essence, the LSTM model defines a transition relationship for the hidden representation $h_{{t}}$ 
through an LSTM cell. Each LSTM cell contains a cell state $c_t$, which serves as a memory and forces the hidden variables $h_t$ to preserve information from the past. 


Specifically, LSTM first generates a candidate cell state $\tilde{c}_t$ by combining $x_t$ and $h_{t-1}$, as: 
\begin{equation}
\small
\begin{aligned}
\tilde{c}_t &= \text{tanh}(W^c_h h_{t-1} + W^c_x x_t).
\end{aligned}
\label{lstm_state}
\end{equation}

LSTM generates a forget gate $f_t$, an input gate  $g_t$, and an output gate $o_t$  via sigmoid function $\sigma(\cdot)$, as:
\begin{equation}
\small
\begin{aligned}
f_t &= \sigma(W^f_h h_{t-1} + W^f_x x_t),\\
g_t &= \sigma(W^g_h h_{t-1} + W^g_x x_t),\\
o_t &= \sigma(W^o_h h_{t-1} + W^o_x x_t).
\end{aligned}
\end{equation}

The forget gate  is used to filter the information inherited from $c^{t-1}$, and the input gate  is used to filter the candidate cell state at $t$. Then we compute the new cell state and the hidden representation as: 
\begin{equation}
\small
\begin{aligned}
c_t &= f_t\otimes c_{t-1}+g_t\otimes\tilde{c}_t,\\
h_t &= o_t\otimes \text{tanh}(c_t),
\end{aligned}
\end{equation}
where $\otimes$ denotes the entry-wise product.


As we wish to conduct regression for continuous values, we generate the predicted temperature $\hat{y_t}$ at each time step $t$ via a linear combination of hidden units, as:
\begin{equation}
\small
    \hat{y_t} = W_y h_t.
    \label{pred}
\end{equation}

We also apply the LSTM model for each depth separately to generate  predictions $\hat{y}_{d,t}$ for every depth $d\in [1,N_d]$ and for every date $t\in [1,T]$. Then given the true observation $y_{d,t}$ for the dates and depths  where the sparse observed data is available, i.e., $S = \{(d,t): y_{d,t} \,\,\text{exists}\}$, our training loss is defined as:
\begin{equation}
\small
    \mathcal{L}_{\text{RNN}} = \sqrt{\frac{1}{|S|}\sum_{(d,t)\in S} (y_{d,t}-\hat{y}_{d,t})^2}.
\end{equation}

It is noteworthy that even if the training loss is only defined on the time steps where the observed data is available, the transition modeling (Eqs.~\ref{lstm_state}-\ref{pred}) can be applied to all the time steps. Hence, the time steps without observed data can still contribute to learning temporal patterns by using their input drivers.

\begin{figure} [!h]
\centering
\label{fig:b}{}
\includegraphics[width=0.6\columnwidth]{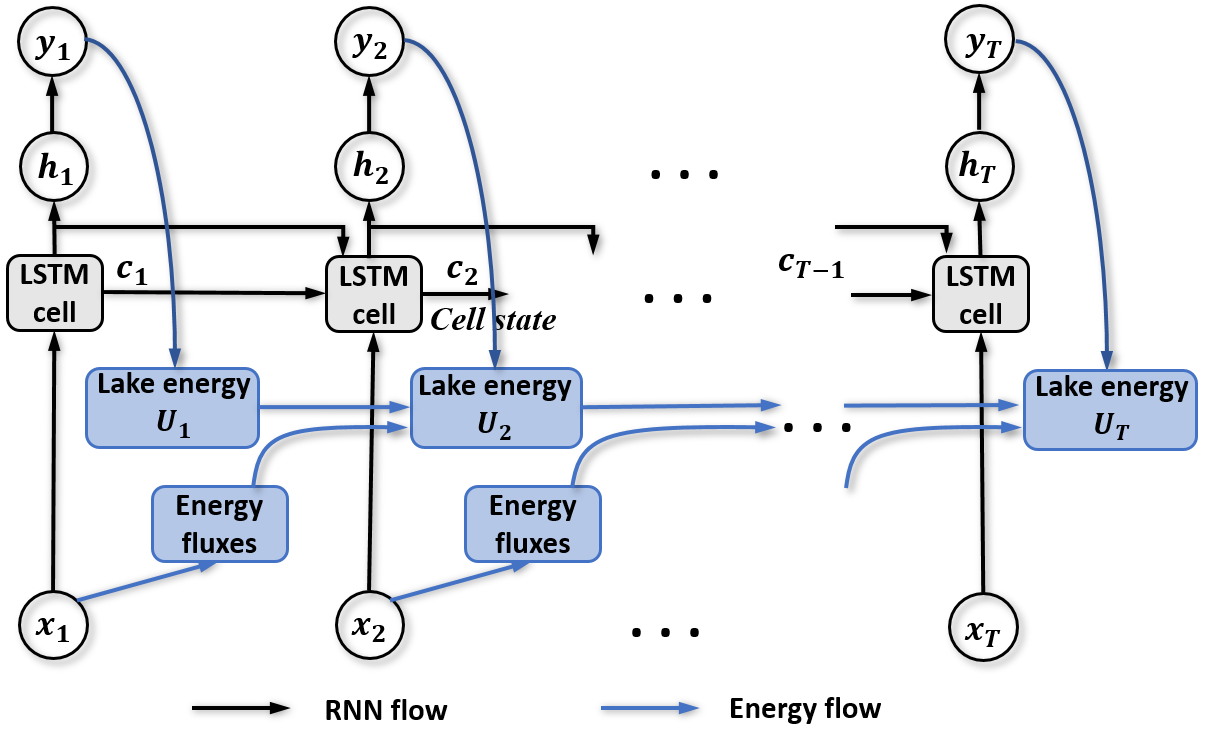}
\caption{The flow of the PGRNN model~\cite{jia2019physics}. The model includes the standard RNN flow (black arrows) and the energy flow (blue arrows) in the recurrent process.}
\label{flow}
\end{figure}

\subsection{Energy conservation over time}
\label{sec:energy-method}

\begin{figure} [!h]
\centering
\label{fig:b}{}
\includegraphics[width=0.4\columnwidth]{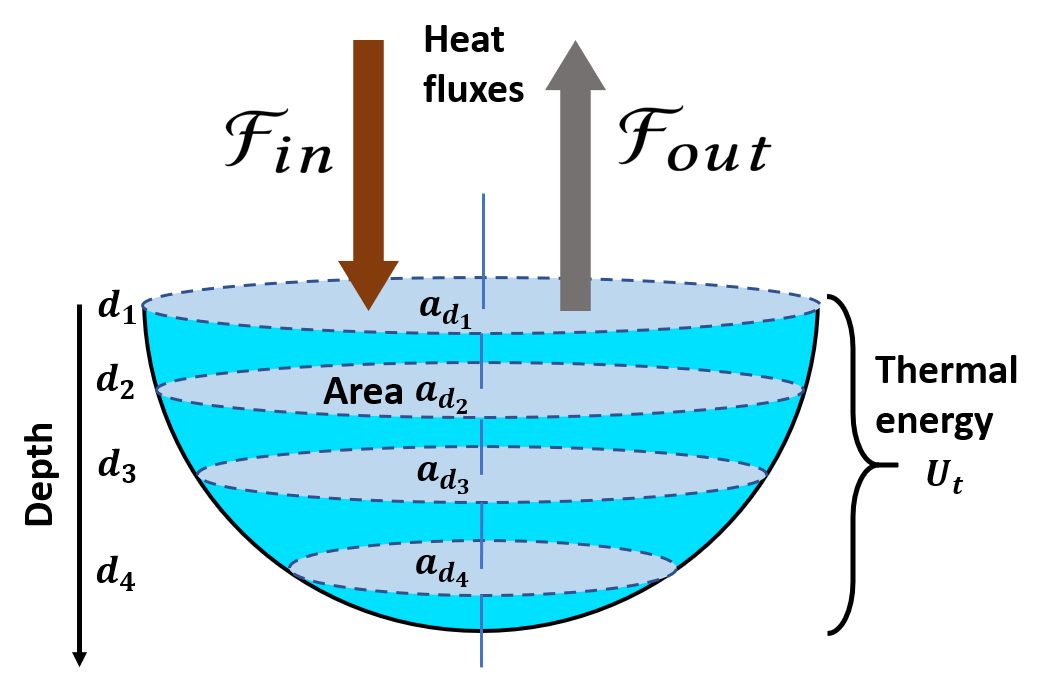}
\caption{The heat energy fluxes and the lake thermal energy that are modeled in PGRNN.}
\label{energy_simp}
\end{figure}

The law of energy conservation states that the change of thermal energy $U_t$ of a lake system over time is equivalent to the net gain of heat energy fluxes, which is the difference between incoming energy fluxes and any energy losses from the lake (see Fig.~\ref{energy_simp}). The explicit modeling of energy conservation is critical for capturing temperature dynamics since a mismatch in losses and gains results in a temperature change. Specifically, more incoming heat fluxes than outgoing heat fluxes will warm the lake, and more outgoing heat fluxes than incoming heat fluxes will cool the lake.

The total thermal energy of the lake at time $t$ can be computed as follows:
\begin{equation}
\small
    U_t = c_w \sum_d a_d y_{d,t} \rho_{d,t} \partial{z_d},
    \label{eq:energy_temp}
\end{equation}
where $y_{d,t}$ is the temperature at depth $d$ at time $t$, $c_w$ the specific heat of water (4186 J kg$^{-1}$\textdegree{}C$^{-1}$), $a_d$ the cross-sectional area of the water column (m$^2$) at depth $d$, $\rho_{d,t}$ the water density (kg/m$^3$) at depth $d$ at time $t$, and $\partial{z_d}$ the thickness (m) of the layer at depth $d$. 
In this work, we simulate water temperature for every 0.5m and thus we set $\partial{z_d}$=0.5. The computation of $U_t$ requires the output of temperature $y_{d,t}$ through a feed-forward process for all the depths, as well as the cross-sectional area $a_d$, which is available as input.

The balance between incoming heat fluxes ($\mathcal{F}_{in}$) and outgoing heat fluxes ($\mathcal{F}_{out}$) results in a change in the thermal energy ($U_t$) of the lake. 
The consistency between lake energy $U_t$ and energy fluxes can be expressed as:
\begin{equation}
\small
\Delta U_t = \mathcal{F}_{in}-\mathcal{F}_{out}
\label{law}
\end{equation}
where $\Delta U_t=U_{t+1}-U_t$. More details about computing heat fluxes are described in the appendix.  
All the involved energy components are in Wm$^{-2}$.  

In Fig.~\ref{flow}, we show the flow of the proposed PGRNN model, which integrates energy conservation flow into the recurrent process. While the recurrent flow in the standard RNN can capture data dependencies across time, the modeling of energy flow ensures that the change of lake environment and predicted temperature conforms to the law of energy conservation. Traditional LSTM models utilize the LSTM cell to implicitly encode useful information at each time step and pass it to the next time step. In contrast, the energy flow in PGRNN explicitly captures the key factor that leads to temperature change in dynamical systems - the heat energy fluxes that are transferred from one time to the next. 
{Standard LSTM model trained from certain years or seasons may not generalize to other years or seasons given that the distributions  of input features and temperature profiles are different in different time periods. However, Eq.~\ref{law} should always hold for data from any time period due to conservation of energy. } Therefore, by complying with the universal law of energy conservation, PGRNN has a better chance at {learning  patterns that are generalizable} to unseen scenarios 
~\cite{read2019process}.




We define the loss function term for energy conservation and combine this with the training objective of standard LSTM model in the following equation:
\begin{equation}
\small
\begin{aligned}
    \mathcal{L} &= \mathcal{L}_{\text{RNN}}+\lambda_{\text{EC}}\mathcal{L}_{\text{EC}},\\
    \mathcal{L}_{\text{EC}}&=\frac{1}{T_{\text{ice-free}}} \sum_{t\in \text{ice-free}} \text{ReLU}(|\Delta U_t-(\mathcal{F}_{in}-\mathcal{F}_{out})|-\tau_{\text{EC}}),
\end{aligned}
\label{combined_1}
\end{equation}
where $T_{\text{ice-free}}$ represents the length of the ice-free period.  Here we consider the energy conservation only for ice-free periods since the lake exhibits drastically different reflectance and energy loss dynamics when covered in ice and snow, and the modeling of ice and snow was considered out of scope for this study. 
We provide more details about how to compute the energy fluxes $\mathcal{F}_{in}$ and $\mathcal{F}_{out}$ from input data in the appendix. 
The value $\tau_{\text{EC}}$ is a threshold for the loss of energy conservation. This threshold is introduced because physical processes can be affected by unknown less important factors which are not included in the model, or by observation errors in the metereological data. The function $\text{ReLU}(\cdot)$ is adopted such that only the difference larger than the threshold is counted towards the penalty. In our implementation, the threshold is set as the largest value of $|\Delta U_t-(\mathcal{F}_{in}-\mathcal{F}_{out})|$ in the GLM model for daily averages.  
The hyper-parameter $\lambda_{\text{EC}}$ controls the balance between the loss of the standard RNN and the energy conservation loss. 
{The model is updated using the back-propagation with the ADAM optimizer~\cite{kingma2014adam}.}

Note that the modeling of energy flow using the procedure described above does not require any input of true labels/observations. According to Eqs.~\ref{eh}-\ref{rout}, the heat fluxes and lake energy are computed using only input drivers and predicted temperature. In light of these observations, we can apply this model for semi-supervised training for lake systems which have only a few labeled data points. 


\subsection{Pre-training using physical simulations}


In real-world environmental systems, observed data is limited. For example, amongst the lakes being studied by USGS, less than 1\% of lakes have 100 or more days of temperature observations and less than 5\% of lakes have 10 or more days of temperature observations~\cite{read2017water}. 
Given their complexity, the RNN-based models trained with limited observed data can lead to poor performance. 
{In addition, ML models often require an initial choice of model parameters before training. Poor initialization can cause models to
anchor in local minimum, which is especially true for deep neural networks. 
If physical knowledge can be used to help inform the initialization of the weights, model
training can accelerated (i.e., require fewer epochs for training) and also need fewer training samples to achieve good performance. }

To address these issues, we propose to pre-train the PGRNN model using the simulated data produced by a generic  GLM (also referred to as uncalibrated GLM) that uses default values for parameters. 
In particular, given the input drivers, we run the generic GLM to predict temperature at every depth and at every day. 
These simulated temperature data from the generic GLM are imperfect but they provide a synthetic realization of physical responses of a lake to a given set of meteorological drivers. Hence, pre-training a neural network using simulations from the generic GLM allows the network to emulate a synthetic but physically realistic phenomena. This process results in a more accurate and physically consistent initialized status for the learning model. When applying the pre-trained model to a real system, we fine-tune the model using true observations. Here our hypothesis is that the pre-trained model is much closer to the optimal solution and thus requires less observed data to train a good quality model.  In our experiments, we show that such pre-trained models can achieve high accuracy given only a few observed data points.

\section{Experiments}
In this section, we conduct extensive evaluations for the proposed method. We first show that the RNN model with LSTM cell can capture the dynamics of lake systems. Then we build the 
RNN$^{EC}$ model by incorporating energy conservation, and demonstrate its effectiveness in  maintaining physical consistency while also reducing prediction error. Moreover, we show that the pre-training method can leverage complex knowledge hidden in a physics-based model. In particular, pre-training the RNN$^{EC}$ model even using the simulated data of a lake that is very different than the target lake (in terms of geometry, clarity and the climate conditions) is able to reduce the number of observations needed to train a good quality model. 



\begin{figure} [!h]
\centering
\label{fig:b}{}
\includegraphics[width=0.5\columnwidth]{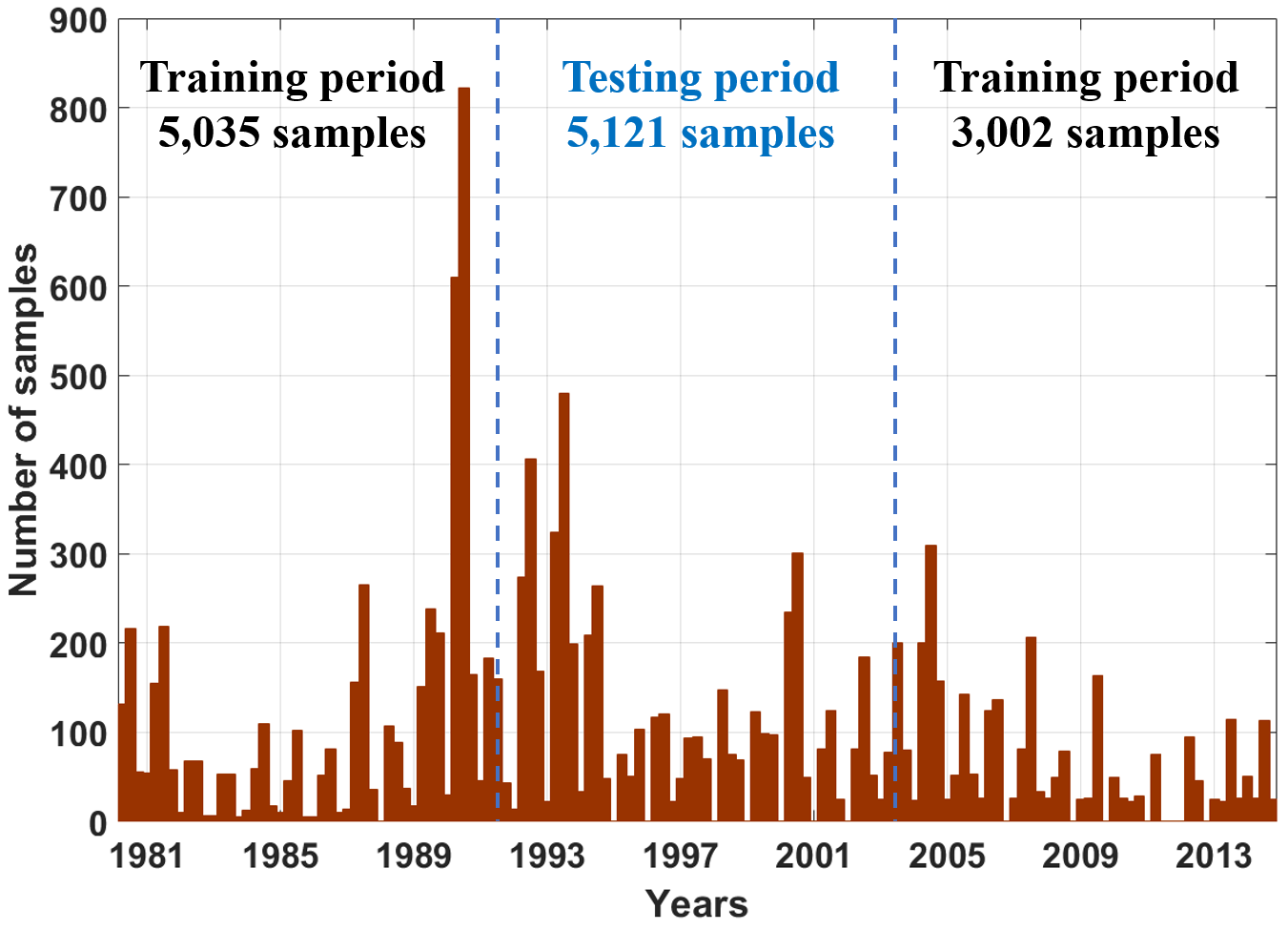}
\caption{The distribution of observed data from April 02, 1980 to December 30, 2014.}
\label{data_dist}
\end{figure}

\subsection{Dataset}
\label{sec:dataset}
Our dataset was collected from Lake Mendota in Wisconsin, USA. This lake system is reasonably large ($\sim$40 km$^2$ in area) {and the lake has a depth of 25 meters. It also}  exhibits large changes in water temperatures in response to seasonal and sub-seasonal weather patterns. {In Table~\ref{temp_range}, we show the range of temperature at different depths.} 
Observations of lake temperature were collected from North Temperate Lakes Long-Term Ecological Research Program~\cite{read2019process}. 
These temperature observations vary in their distribution across depths and time. There are certain days when observations are available at multiple depths while only a few or no observations are available on some other days.

\begin{table}[!h]
\small
\newcommand{\tabincell}[2]{\begin{tabular}{@{}#1@{}}#2\end{tabular}}
\centering
\caption{The minimum, maximum and mean temperature ($^\circ$C) at surface, 12 m and 24 m depths in different seasons.  }
\begin{tabular}{l|ccc|ccc|ccc|ccc}
\hline
 & &Spring& & &Summer& & &Fall&  & &Winter& \\ 
Depth& min&max&mean&min&max&mean&min&max&mean&min&max&mean \\\hline 
0 m (surface)&0.1&23.2&7.8 &6.3&29.2&20.9&6.0&28.6&18.7&0.1&14.3 &4.8  \\
12 m &0.7&14.9&6.5 &6.3&21.9&14.2&7.4&22.5&16.2&0.4&14.3 &5.1 \\
24 m &0.7&12.0&6.2 & 5.9&15.0&10.5& 2.0&15.8&11.4&1.3&12.4&5.7 \\\hline
\end{tabular}
\label{temp_range}
\end{table}

The input drivers that describe prevailing meteorological conditions are available on a continuous daily basis from April 02, 1980 to December 30, 2014 (see~\cite{read2019process} for data access and description of inputs)  
Specifically, we used a set of seven drivers as input variables, which include short-wave and long-wave radiation, air temperature, relative humidity, wind speed, frozen and snowing indicators. In contrast, observed data for training and testing the models is not uniform, as measurements were made at varying temporal and spatial (depth) resolutions. In total, 13,158 observations were used for the study period, as shown in Fig.~\ref{data_dist}.

We use the observed data from  April 02, 1980 to October 31, 1991 and the data from June 01, 2003 to December 30, 2014 as training data (in total 8,037 observations). Then we applied the trained model to predict the temperature at different depths for the period from November 01, 1991 to  May 31, 2003 (in total 5,121 observations).

\subsection{Model setup}
We implement the proposed method using Tensorflow with Tesla P100 GPU. The recurrent modeling structure uses 21 hidden units. The threshold value $\tau_{\text{EC}}$  is set as 24 {W/m$^2$}, which is equivalent to the largest value of $|\Delta U_t-(\mathcal{F}_{in}-\mathcal{F}_{out})|$ in the GLM model for daily averages. The hyper-parameter $\lambda_{EC}$ is set to 0.01. The value of $\lambda_{EC}$ is selected to balance the supervised training loss and the conservation of energy. A smaller value of $\lambda_{EC}$ results in a lower training loss at the expense of conservation of  energy, and vice versa. Note that, when $\lambda_{EC}$>0 (and thus energy conservation is part of the loss function), then the model has a better chance at learning general patterns that can reduce test error. (compared with the test error using $\lambda_{EC}$=0).  
Also note that the energy conservation term is not fully 
accurate since certain minor physical processes are not captured by the energy conservation loss. Hence, a much larger value of $\lambda_{EC}$ can also results in sub-optimal performance by enforcing the model to conform to approximate physical relationships.    The model is trained with the learning rate of 0.005. 

{We also pre-train the model using the simulation data provided by a generic GLM model. The effectiveness of the pre-training strategy is discussed in Section~\ref{sec:pretrain}. When we show the predictive performance of the GLM model, we refer to the GLM model which has its parameters (e.g., geometry, water clarity, salinity) calibrated using the same amount of data that is used to train ML models. We use GLM-gnr to represent a GLM model that uses generic values for the parameters.}

\subsection{Performance: prediction accuracy and energy consistency}


First, we aim to evaluate how energy conservation helps improve the prediction accuracy and maintain the energy consistency. In our experiments, we use RNN to represent the RNN model with the LSTM cell, and use the RNN$^{EC}$ to represent the LSTM-RNN networks after incorporating energy conservation to the entire study period. We assess the performance of each model based on their prediction accuracy (see Section~\ref{sec:prediction_accuracy}) and the physical consistency (see Section~\ref{sec:energy_consistency}).  
Some sensitivity tests regarding to hyper-parameters can be found in our previous work~\cite{jia2019physics}.

\subsubsection{Prediction accuracy}
\label{sec:prediction_accuracy}

Here we compare RNN, RNN$^{EC}$, and GLM in terms of their prediction RMSE~\footnote{Here we do not include the basic neural network and the standard RNN model (without LSTM cell) since the basic neural network produces an RMSE of 1.88  and the standard RNN produces an RMSE of 1.60 using 100\% observed data, which is far higher than the models under discussion. }. 
To test whether each model can perform well using reduced observed data. We randomly select a different proportion of data from the training period. For example, to select 20\% of training data, we remove every observation in our training period with 0.8 probability. The test data stays the same regardless of training data selection. We repeat each test 10 times and report the mean RMSE and standard deviation. 

From Table~\ref{prediction}, we have several observations: 1) RNN$^{EC}$ consistently outperforms RNN. The gap is especially obvious when using smaller subsets of observed data (e.g., 0.2\% or 2\% data). However, given plenty of observed data, the RNN model can achieve the similar performance with the RNN$^{EC}$ model. 2) Both RNN and RNN$^{EC}$ can get close to their best performance using over 20\% observed data.  3) RNN$^{EC}$ using 20\% observed data 
outperforms fully calibrated GLM (using 100\% observed data).

\begin{table}[!h]
\newcommand{\tabincell}[2]{\begin{tabular}{@{}#1@{}}#2\end{tabular}}
\centering
\caption{Performance {(as measured by RMSE in degrees Celsius)} of RNN, RNN$^{EC}$ and GLM with access to  different amount of observed data. 
}
\begin{tabular}{l|ccccc}
\hline
\textbf{Method}& \textbf{0\%} & \textbf{0.2\%} & \textbf{2\%}  & \textbf{20\%} & \textbf{100\%} \\ \hline 
GLM & 2.950($\pm$NA)& 2.616($\pm$0.499)& 2.422($\pm$0.423)& 2.318($\pm$0.368) &1.836($\pm$NA)\\ 
RNN  & - &4.615($\pm$0.173) &2.311($\pm$0.240) &   1.531($\pm$0.083) &   1.489($\pm$0.091) \\ 
RNN$^{EC}$  & - &4.107($\pm$0.181) & 2.149($\pm$0.163) & 1.489($\pm$0.115) &   1.471($\pm$0.077) \\ \hline
\end{tabular}
\label{prediction}
\end{table}

\subsubsection{Energy consistency}
\label{sec:energy_consistency}
\begin{figure} [!h]
\centering
\label{fig:b}{}
\includegraphics[width=0.35\columnwidth]{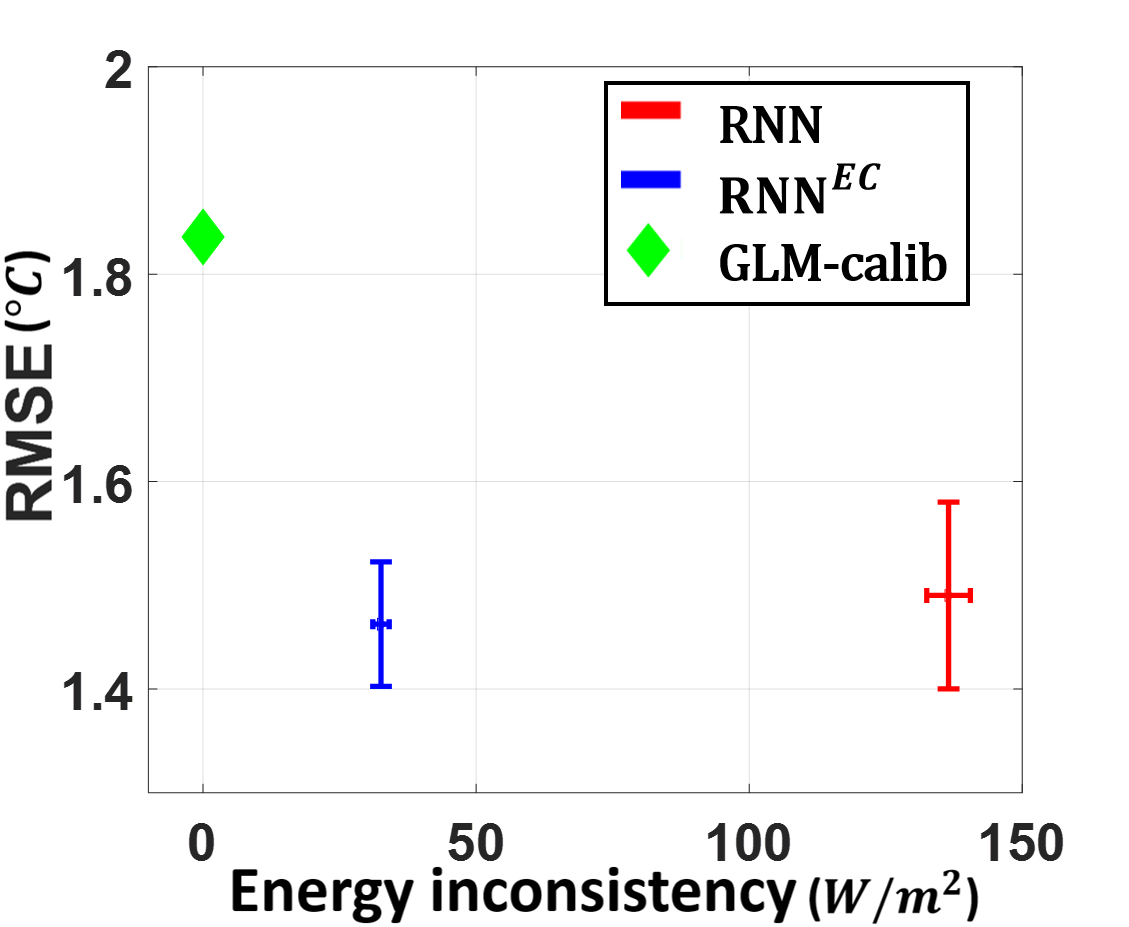} 
\caption{The performance of RNN, RNN$^{EC}$ and calibrated GLM by RMSE and energy inconsistency.}
\label{ec_res}
\end{figure}

To visualize how RNN$^{EC}$ contributes to a physically consistent solution, we wish to verify whether the gap between incoming and outgoing heat energy fluxes matches the lake energy change over time. Specifically, we train RNN and RNN$^{EC}$ using observed data from the first ten years.    
{We measure the difference between the lake's energy change (left-hand side of Eq.~\ref{law}) and the net gain of energy fluxes (right-hand side of Eq.~\ref{law}) and then represent the average difference as the energy inconsistency.}  In Fig.~\ref{ec_res}, we show the RMSE and the energy inconsistency of RNN, RNN$^{EC}$ and the calibrated GLM model in the entire test period. Here each model is trained using 100\% observed data (the last column in Table~\ref{prediction}).



We observe that RNN$^{EC}$ produces a better match between energy fluxes and lake energy change while RNN leads to a large energy inconsistency. 
This confirms that the addition of energy conservation term in the loss function used for RNN$^{EC}$ during its training period results in a model that helps preserve energy conservation in the test data.  Note that {RNN$^{EC}$ still has a larger energy inconsistency than the calibrated GLM}.   RNN$^{EC}$ can obtain {a smaller energy inconsistency} by simply using a larger value of $\lambda_{EC}$ during the training phase.  However, the energy conservation formula used in Eqs.~\ref{law} and~\ref{law_detail} (in Appendix) captures only a subset of physical processes and ignores certain minor processes that can be challenging to be precisely modeled~\cite{read2019process}, and thus strict compliance to the simplified energy conservation term used in the loss function of RNN$^{EC}$ can reduce the prediction accuracy in unseen data.  Finally, from Figure~\ref{ec_res} (and also from Table~\ref{prediction}), we can see that RNN$^{EC}$ has even lower RMSE than RNN (which focuses only on reducing RMSE during the training phase).  This shows that a more physically realistic model can also be more generalizable.

\subsection{Leveraging the knowledge hidden in physics-based model via pre-training}
\label{sec:pretrain}

Here we show the power of pre-training to improve prediction accuracy 
of the model even with small amounts of training data.  
A basic premise of pre-training our models is that GLM simulations, though imperfect, provide a synthetic realization of physical responses of a lake to a given set of meteorological drivers. Hence, pre-training a neural network using GLM simulations allows the network to emulate a synthetic realization of physical   phenomena. Our hypothesis is that such a pre-trained model requires fewer labeled samples to achieve good generalization performance, even if the GLM simulations do not match with the observations. To test this hypothesis, we conduct an experiment where we generate GLM simulations with input drivers from Lake Mendota. These simulations have been created using a GLM model with generic parameter values that are not calibrated for Lake Mendota, resulting in large errors in modeled temperature profiles with respect to the real observations on Lake Mendota (RMSE=2.950). Nevertheless, these simulated data are physically consistent and by using them for pre-training, we can demonstrate the power of our ML models to work with limited observed data while leveraging complex physical knowledge inherent in the physical models. {In particular, we generate simulation data for every single date from April 02, 1980 to December 30, 2014 and for every depth. We use all the simulation data to pre-train the RNN$^{EC}$ model by minimizing the loss function defined in Eq.~\ref{combined_1}. }

We fine-tune the pre-trained models with different amounts of observed data and report the performance  in Table~\ref{pre-training}. We use the notation, RNN$^{EC,p}$, to refer to the RNN model with energy conservation that is first pretrained using simulation data during 1981-2013 and then gets fine-tuned using observed data from the training period.  
The comparison between RNN$^{EC}$ and RNN$^{EC,p}$ shows that the pre-training can significantly improve the performance. The improvement is 
relatively much larger given a small amount of observed data. 
For example, even with 0.2\% of observed data (16 observations) RNN$^{EC,p}$ achieves RMSE of 2.056, which is much smaller than that obtained by RNN or RNN$^{EC}$ when using ten times the amount of observed data. 
Moreover, we find that the training RNN and RNN$^{EC}$ model commonly takes 150-200 epochs to converge while the training for RNN$_p$ and RNN$^{EC,p}$ only takes 30-50 epochs to converge. The improvements in these aspects demonstrate that pre-training can indeed provide a better initialized state for learning a good quality model.

\begin{figure} [!h]
\centering
\label{fig:b}{}
\includegraphics[width=0.8\columnwidth]{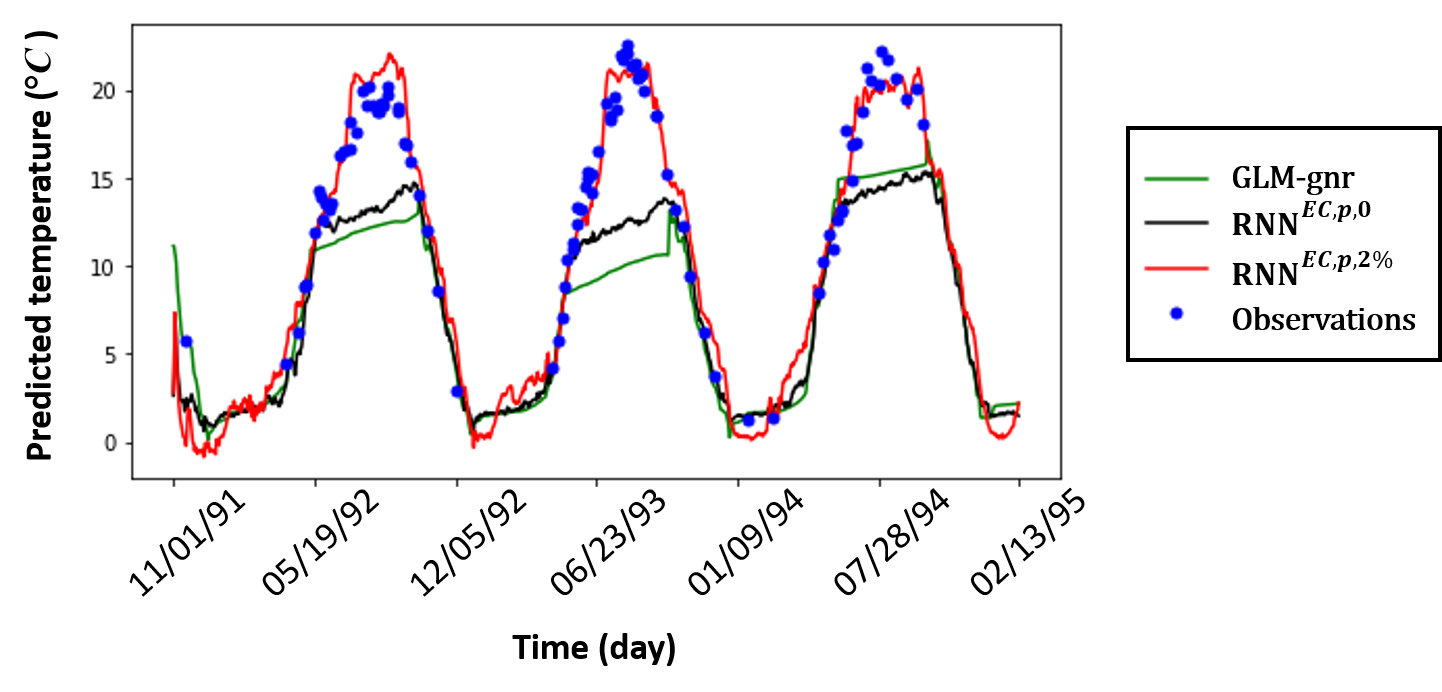}
\caption{The predictions by the generic GLM (GLM-gnr), RNN$^{EC,p,0}$ (the pretrained RNN$^{EC}$ model), RNN$^{EC,p,2\%}$ (the pretrained RNN$^{EC}$ model which then gets fine-tuned using 2\% training observations) at 10m depth from November 01, 1991 to February 13, 1995. }
\label{fine-tuning-help}
\end{figure}

Now we wish to better understand how the fine-tuning improves the performance using only limited observations.  In Fig.~\ref{fine-tuning-help}, we show the predictions at 10 m depth by the generic GLM (i.e., GLM-gnr), the pretrained RNN$^{EC}$ without fine-tuning (i.e., RNN$^{EC,p,0}$), and the pretrained RNN$^{EC}$ using 2\% data for fine-tuning (i.e., RNN$^{EC,p,2\%}$). We include the distribution of the randomly selected 2\% training data in the appendix. We have following observations: 1) The generic GLM results in a large bias with true observations. 2) RNN$^{EC,p,0}$ has similar predictions with the generic GLM since RNN$^{EC,p,0}$ is pretrained to emulate the generic GLM. Note that RNN$^{EC,p,0}$ has roughly captured temperature dynamics even without using any observed data. 3) After fine-tuning using just 2\% observed data, the RNN$^{EC,p,2\%}$ very well closes the gap between RNN$^{EC,p,0}$ and true observations.

\begin{table}[!h]
\small
\newcommand{\tabincell}[2]{\begin{tabular}{@{}#1@{}}#2\end{tabular}}
\centering
\caption{Performance of the pre-trained model (RNN$^{EC,p}$) after they are fine-tuned with access to different amount of observed data. We include the performance of GLM, RNN and RNN$^{EC}$ from Table 1 here to allow for an easier comparison. }
\begin{tabular}{l|ccccc}
\hline
\textbf{Method} & \textbf{0\%} & \textbf{0.2\%} & \textbf{2\%}  & \textbf{20\%} & \textbf{100\%} \\ \hline 
GLM & 2.950& 2.616($\pm$0.499) & 2.422($\pm$0.423)& 2.318($\pm$0.368) &1.836($\pm$NA)\\ 
RNN & - & 4.615($\pm$0.173) &2.311($\pm$0.240) &   1.531($\pm$0.083) &   1.489($\pm$0.091) \\ 
RNN$^{EC}$  & - &4.107($\pm$0.181) & 2.149($\pm$0.163) & 1.489($\pm$0.115) &   1.471($\pm$0.077) \\ 
RNN$^{EC,p}$& 2.455($\pm$0.169)& 2.056($\pm$0.180)& 1.590($\pm$0.162)   &   1.402($\pm$0.106)    &1.380($\pm$0.078)\\ \hline  
\end{tabular}
\label{pre-training}
\end{table}

\subsection{The RMSE profile across depths and seasons}



Here we further analyze the prediction results to understand the limitations of physics-based GLM models and how our proposed method can overcome these limitations. Specifically, we conduct analysis from two different perspectives - across depths and across seasons. Each one will provide unique insights on the underlying difference between GLM and the proposed method in modeling lake temperature dynamics.


\subsubsection{Error across depths:} 
\begin{figure} [!h]
\centering
\includegraphics[width=0.45\columnwidth]{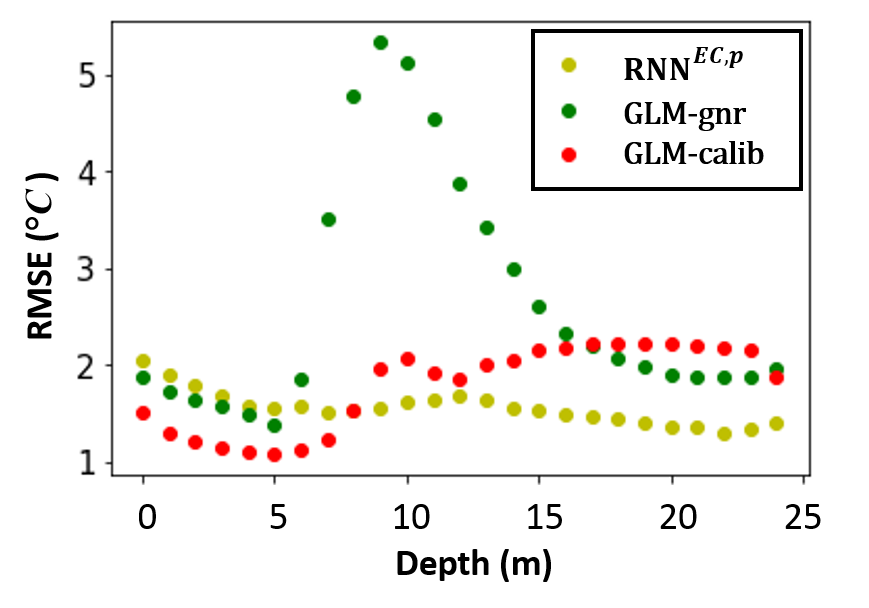}
\caption{ The testing RMSE error at different depths. The errors are calculated only at the depths where more than 50 observed data are available. 
RNN$^{EC,p}$ represent the 
RNN$^{EC}$ model that is pre-trained with simulated data and then fine-tuned by 100\% observed data. GLM-{gnr} and GLM-{calib} represent the generic GLM and the fully calibrated GLM (using 100\% observed data), respectively. }
\label{err_depth_overall}
\end{figure}

In Fig.~\ref{err_depth_overall}, we show the error of RNN$^{EC,p}$ models (pre-trained and fine-tuned with 
100\% data) and the GLM models (generic GLM and calibrated GLM using 100\% data) across different depths.

It can be seen at the shallow depth levels (< 6 m),  RNN$^{EC,p}$ model achieves  similar performance with the generic GLM, but has larger errors than the calibrated GLM. This is because a single RNN$^{EC,p}$ model is trained to optimize the performance  across all the depths. If we separately train an RNN$^{EC,p}$ only for shallow depths, the performance can be close to the calibrated GLM.

The generic GLM model has much larger errors than RNN$^{EC,p}$ at depths larger than 6 m, especially at intermediate depths (i.e., between 6 m - 16 m). 
The reason for such depth-dependent differences between GLM and RNN$^{EC,p}$ is because GLM includes complex processes to model the dynamics of thermal stratification, which includes the density-based separation of the surface and bottom waters. Specifically, the GLM is designed to capture the location of this temperature transition and strength of the gradient. However, predicting the dynamics of stratification from the basis of the underlying processes is very challenging for any model, including the GLM~\cite{hipsey2019general},  
and thus we can observe an increase in errors of the generic GLM model at depth layers below 6 m.

The calibrated GLM has much smaller errors than the generic GLM at middle depths. This shows that the generic GLM model simulates complex processes that cannot be easily generalized to specific lake systems without calibration. 
After GLM is calibrated using true observations, it can better locate the temperature transition in this specific lake and consequently reduce the errors in the middle depths.  
Note that  the calibrated GLM 
still has larger errors  compared to RNN$^{EC,p}$ at lower depths.
This is potentially the result of challenges from a physics-based formulation of stratification dynamics. 
In contrast, the ML models approach the problem of prediction without making any assumptions of the stratification processes, and are able to perform much better at intermediate and lower depths by learning patterns from the training data. 

\subsubsection{Error across seasons:} 
\begin{figure} [!h]
\centering
\label{fig:b}{}
\includegraphics[width=0.4\columnwidth]{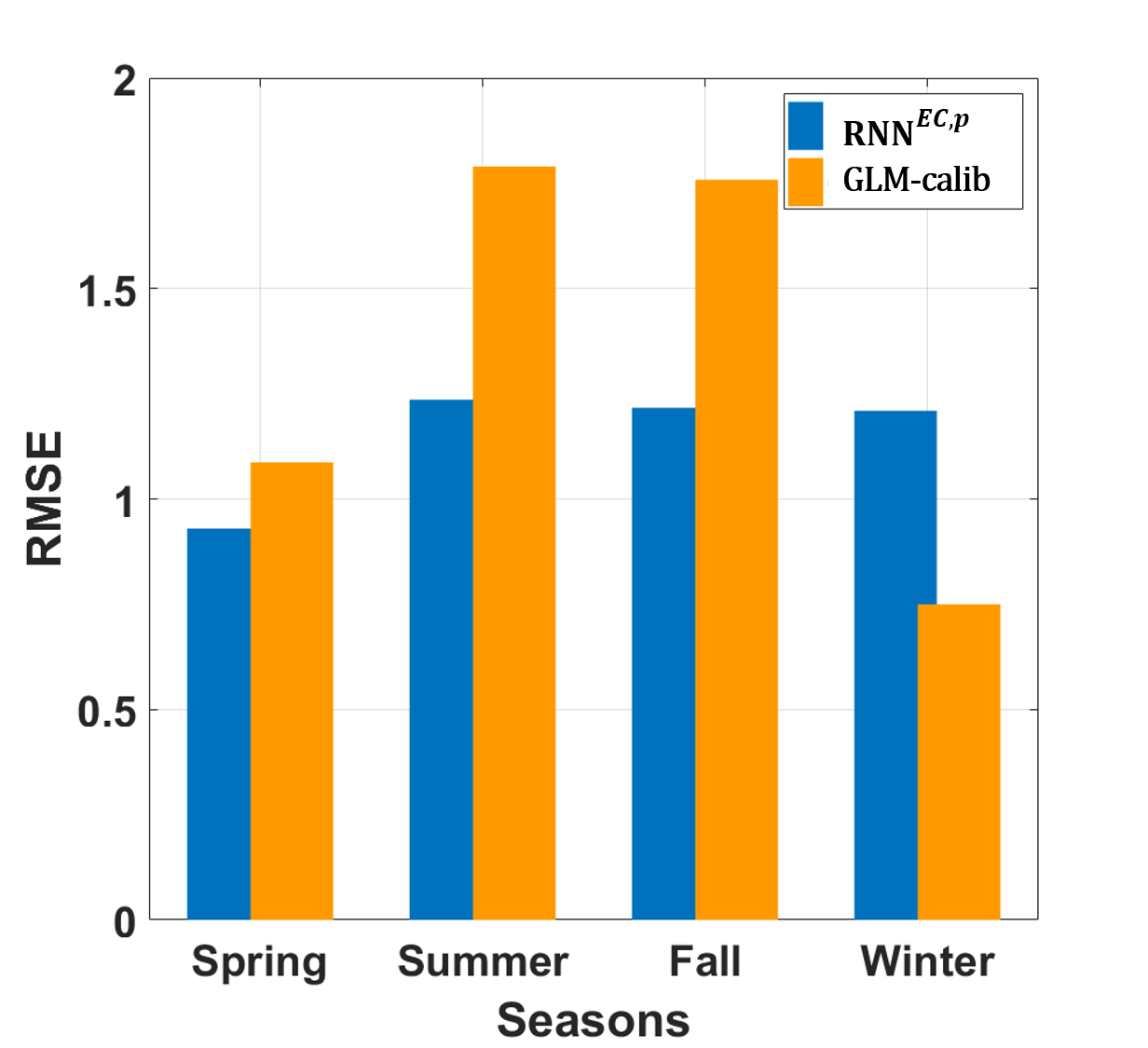}
\caption{ The overall error of RNN$^{EC}$ and calibrated GLM model in different seasons.}
\label{err_seasons}
\end{figure}

We show the overall error in each season in Fig.~\ref{err_seasons}. We can observe that in spring RNN$^{EC,p}$ and calibrated GLM have similar errors, while in summer and fall RNN$^{EC,p}$ outperforms calibrated GLM by a considerable margin, with calibrated GLM offering  improvement over RNN$^{EC,p}$ during the winter season. 
This implies a bias by GLM in modeling certain physical processes that are active during warmer seasons.

\begin{figure} [!h]
\centering
\label{fig:b}{}
\includegraphics[width=0.8\columnwidth]{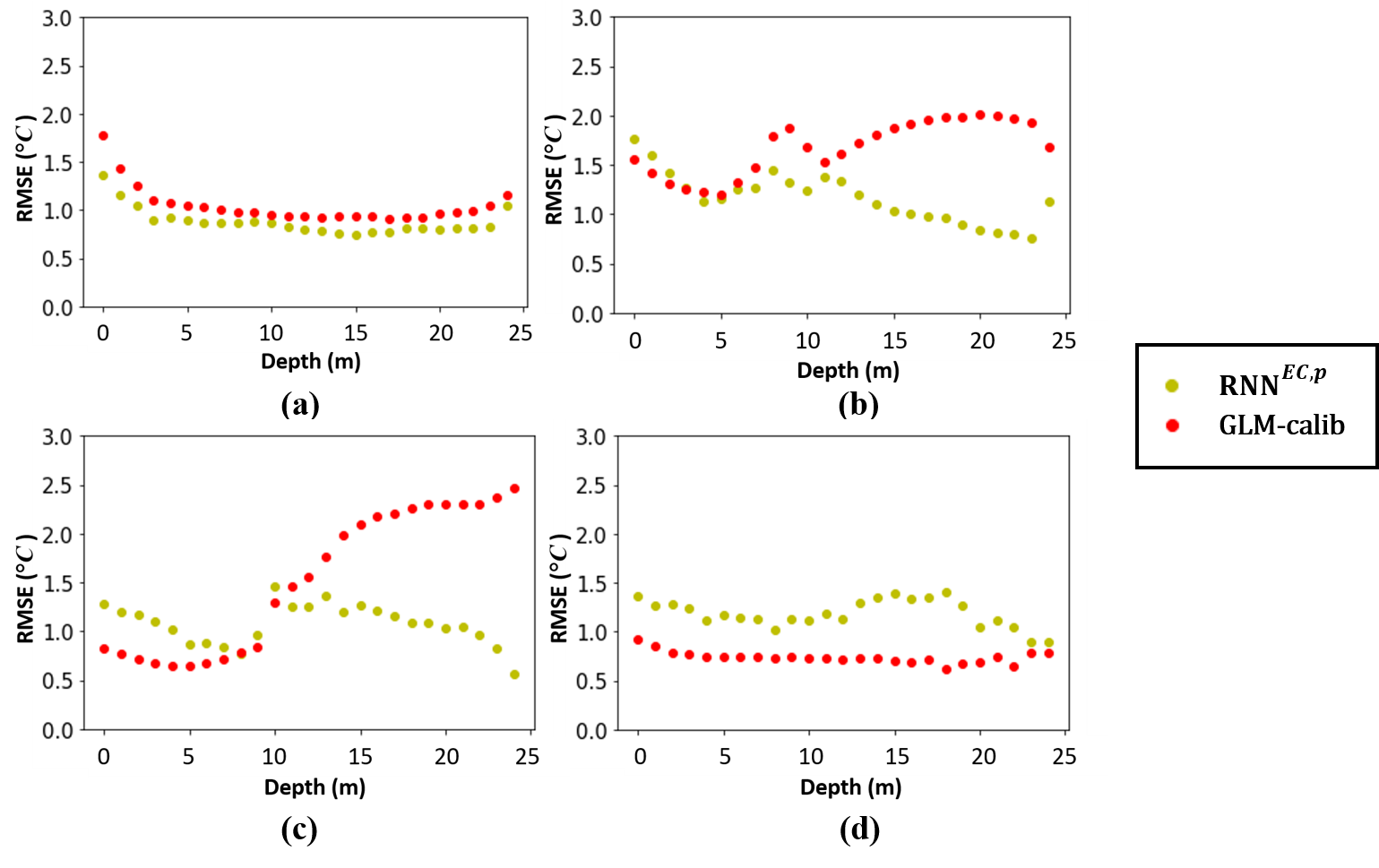}
\caption{ The error-depth relationship in (a) spring, (b) summer, (c) fall, and (d) winter. }
\label{err_depth_seasons}
\end{figure}

To better understand the difference between our proposed method and GLM  across seasons, we separately plot the error-depth relation for different seasons  (see Fig.~\ref{err_depth_seasons}). 
We can observe the error-depth profile in summer and fall are similar to that in Fig.~\ref{err_depth_overall}. The difference between RNN$^{EC,p}$ and calibrated GLM performance is 
especially worse in summer and fall because these two seasons are dominated by a stronger stratification and/or rapid changes in stratification as the lake cools. The influence of stratification on model performance in the spring and winter period 
is weaker compared to summer and fall. 
Hence, the difficulty in modeling stratification in addition to the increased range of temperatures are likely responsible for GLM's worse performance when compared to RNN$^{EC,p}$ in warmer seasons.








\subsection{Pre-trained ML model vs. its teacher}

\begin{figure} [!h]
\centering
\label{fig:b}{}
\includegraphics[width=0.4\columnwidth]{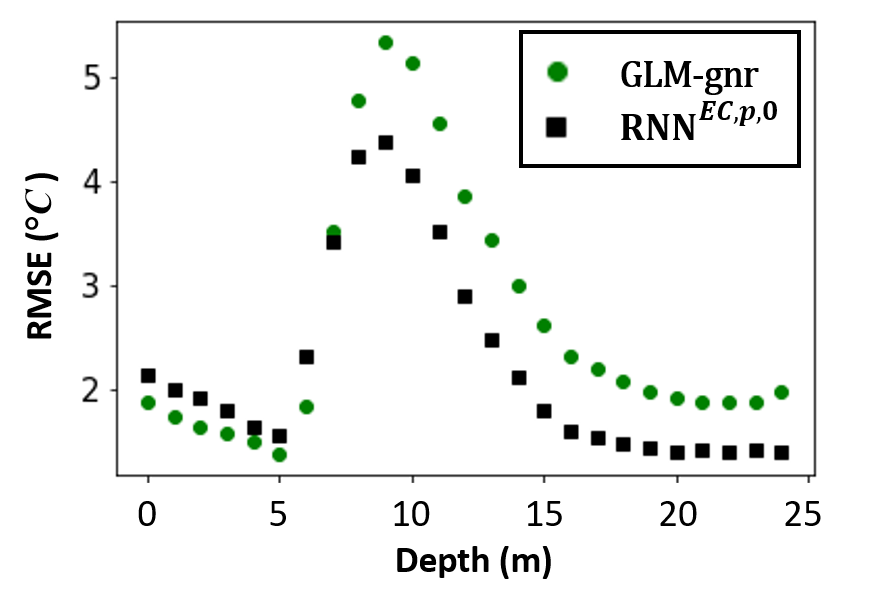}
\caption{The error of GLM and pre-trained model (i.e., RNN$^{EC,p,0}$)  at different depths. No data is used for refinement. 
}
\label{pre-trained_err}
\end{figure}

As observed from Table~\ref{pre-training}, the performance of the pre-trained RNN-based models with no fine-tuning is better than the accuracy of the outputs from the generic GLM model (RMSE=2.950) based on which 
RNN$^{EC}$ is pre-trained. 
GLM tracks temperature at various depth layers that grow and shrink, split, or combine based on prevailing conditions (this is referred to as a Lagrangian layer model, since vertical layers are not fixed in time). As adjacent layers split or combine, prediction artifacts that are not representative of the real-world lake system are introduced, which often result in additional variability at lower depths. The resulting temperature variability can be overly sensitive for Lake Mendota and can increase GLM error. In contrast, the pre-trained  RNN$^{EC}$ as an  imperfect emulator of GLM does not fully capture such complexity, and instead predicts smoother and often more accurate temperature dynamics compared to  the simulated data. 
To verify that GLM can introduce unnecessary variability or temperature change artifacts at lower depths that are comparatively muted in the pre-trained model, in Fig.~\ref{pre-trained_err} we show the error profile of GLM and the pre-trained model at different depths when no observations are used for refinement, i.e., the RNN$^{EC,p,0}$ model.  
We can observe that the pre-trained 
RNN$^{EC,p,0}$ model and GLM achieve similar performance around the surface but the pre-trained RNN$^{EC,p,0}$ has much lower RMSE than the GLM model at lower depths.

To better illustrate this, we pre-train 
RNN$^{EC,p,0}$ using data from different depth layers - surface (0 m) and 9 m. 
Then we measure the error of each model with respect to GLM simulated data and true observation data at the same depth where the model is trained (Table~\ref{analysis_1}). We can observe that the error to GLM outputs is much higher at 9 m than at the surface.  This shows that the ML models cannot fully mimic the complexity of  GLM at lower depths. However, since these complex processes are not necessarily good representations of Lake Mendota temperature dynamics, the ML models achieve lower RMSE to true observations compared to GLM (
4.752 by RNN$^{EC,p,0}$, and 5.333 by GLM) at 9 m by learning a simpler temporal process that is closer to reality.

\begin{table}[!h]
\small
\newcommand{\tabincell}[2]{\begin{tabular}{@{}#1@{}}#2\end{tabular}}
\centering
\caption{The error of GLM and pre-trained models with respect to GLM simulations and observation data at different depths.}
\begin{tabular}{l|cc|cc}
\hline
& \multicolumn{2}{c|}{\textbf{Surface}} & \multicolumn{2}{c}{\textbf{9 m}}  \\ 
\textbf{Method} & Simulation error & Observation error &  Simulation error & Observation error \\ \hline 
GLM-gnr & - & 1.875 & - &  5.333 \\ 
RNN$^{EC,p,0}$ & 0.854 & 1.932 & 1.498 & 4.752\\ \hline 
\end{tabular}
\label{analysis_1}
\end{table}




\subsection{Ability to pre-train using lakes that are very different {than} target lake}

In practice, the GLM model may not have access to true values of parameters (e.g., lake geometry, water clarity and climate conditions), and therefore can only generate simulations based on default and inaccurate assumptions of parameters that influence lake temperature dynamics. Here we show the power of pre-training using simulated data from a physics-based model built on different lake geometries, lake clarity, and climate conditions.  Our assumption is that the simulations by physics-based models still represent physical responses that strictly follow known physical laws. Hence, the pre-trained model should be able to capture these physical relationships and reach a physically-consistent initialized state. In our experiment, we will show that the pre-training with even a wrong set of lake parameters or with weather drivers very different from the target lake can still significantly reduce the amount of observations required to train a good quality model.



Specifically, we pre-train 
RNN$^{EC}$ using the simulated data by GLM based on specific conditions (geometry, clarity, and climate conditions). Then we will verify whether {these} pre-trained models still have superior performance after they are fine-tuned with a small amount of observations. 


\noindent{\underline{Lake geometry:}} 

\begin{figure} [!h]
\centering
\label{fig:b}{}
\includegraphics[width=0.3\columnwidth]{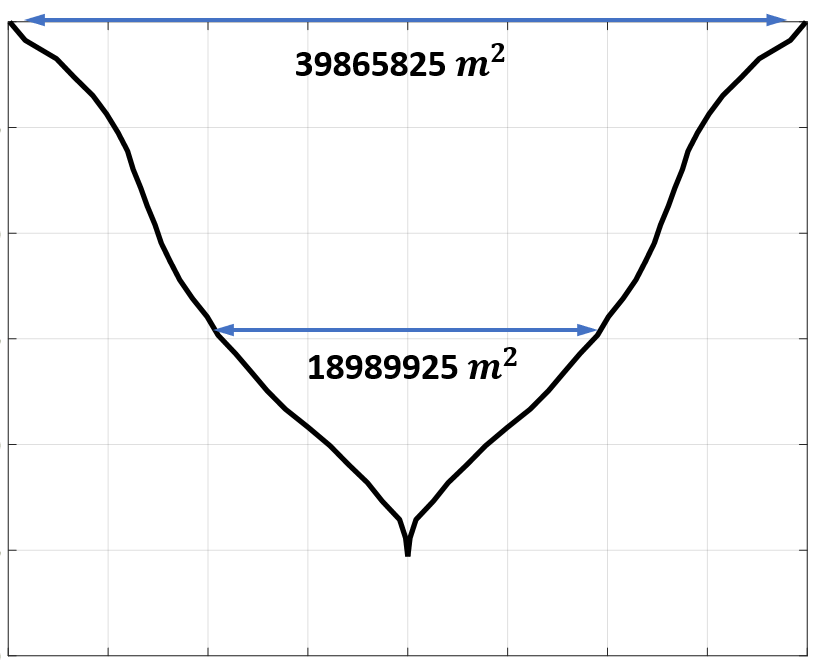}
\caption{The geometry of Lake Mendota. }
\label{fig:geometry}
\end{figure}

We generate GLM simulations for three synthetic lakes with three different lake geometric structure: cone, barrel, and martini. The cone shape is closer to the true geometry of Lake Mendota (see Fig.~\ref{fig:geometry}) while both barrel and martini are very different to the true geometry.  
We first conduct pre-training using the GLM outputs based on each geometric structure. Then we conduct fine-tuning using true observations. The performance is shown in Table~\ref{geometry}.

It can be seen that when adapted to Lake Mendota, the learned model from the cone shape works well even with no observed data.  
In contrast, the models learned from the barrel and martini shapes have a much larger error when directly applied to Lake Mendota. However, these errors are significantly reduced after fine-tuning with only 2\% data. This shows that the model learned from a specific geometric structure can  also capture certain temporal patterns that are physically consistent and applicable to the target system.  

\begin{table}[!h]
\small
\newcommand{\tabincell}[2]{\begin{tabular}{@{}#1@{}}#2\end{tabular}}
\centering
\caption{Performance of pre-trained models from different geometric structures (cone, martini and barrel) after they are fine-tuned with different amount of observed data from Lake Mendota.  }
\begin{tabular}{l|ccccc}
\hline
\textbf{Method} & \textbf{0\%} & \textbf{0.2\%} & \textbf{2\%}  & \textbf{20\%} & \textbf{100\%} \\ \hline 
RNN$^{EC}$  & - &4.107($\pm$0.181) & 2.149($\pm$0.163) & 1.489($\pm$0.115) &   1.471($\pm$0.077) \\ 
RNN$^{EC,p}$& 2.455($\pm$0.169)& 2.056($\pm$0.180)& 1.590($\pm$0.162)   &   1.402($\pm$0.106)    &1.380($\pm$0.078)\\ 
RNN$^{EC,p\text{(cone)}}$& 2.469($\pm$0.168) & 2.056($\pm$0.184) & 1.595($\pm$0.097) & 1.452($\pm$0.113) & 1.374($\pm$0.074) \\ 
RNN$^{EC,p\text{(barrel)}}$& 3.239($\pm$0.098) & 2.060($\pm$0.144)& 1.617($\pm$0.090) & 1.401($\pm$0.098) & 1.383($\pm$0.078)\\ 
RNN$^{EC,p\text{(martini)}}$& 5.340($\pm$0.110) & 3.033($\pm$0.104) & 2.216($\pm$0.141)  & 1.485($\pm$0.092) & 1.459($\pm$0.059)\\ \hline 
\end{tabular}
\label{geometry}
\end{table}

When comparing the performance of different pre-trained geometric structures, we notice that the model pre-trained with the martini shape has a much larger error (RMSE 5.340) than the other two geometric shapes and the cone shape has the smallest error (see the first column in Table~\ref{geometry}). This result agrees with the assumption that the cone shape is closer to the true geometry of Lake Mendota. 
Consequently, the GLM simulations using the cone shape should be closer to reality and the GLM simulations in martini shape should be far away from true observations. We verify this by measuring the RMSE of the GLM simulations with respect to true observation data: \{cone simulation=2.792, martini simulation=5.950, barrel simmulation=3.864\}. Even though the GLM simulations can have large errors when assuming the wrong geometric structure,  the pre-trained models obtain lower errors than their teacher  (see the first column in Table~\ref{geometry}: \{cone 2.469, martini 5.340, barrel 3.239\}). This shows that the machine learning models are less sensitive to the change of geometric structure. Moreover, even though the models pre-trained using the wrong geometric structure have relatively large errors after pre-training, they can quickly recover to reasonable performance when fine-tuned with small amount of true observations data (e.g., 2\% data).


\noindent{\underline{Lake clarity:}} 

Similarly, we generate GLM simulations for three synthetic lakes with different levels of clarity: normal (Kw=0.45), dark (Kw=1.20) and clear (Kw=0.25). Here we fix the lake geometry as a cone shape. The clarity level affects the penetration rate of radiation into the deeper water. We wish to verify how a model learned from a different clarity level can be fine-tuned to fit Lake Mendota. The performance is shown in Table~\ref{clarity}.

We can observe that even if the Lake Mendota has the clarity level Kw close to the normal (Kw=0.45) level, the  model pre-trained from both ``dark" clarity and ``clear" clarity can be well adapted to lake Mendota after fine-tuning. We also note that the performance of fine-tuned models 
from different clarity levels are similar given even 0.2\% observations. 
This shows that the clarity level has less impact than lake geometry on learning an accurate predictive model for lake systems. 

\begin{table}[!h]
\small
\newcommand{\tabincell}[2]{\begin{tabular}{@{}#1@{}}#2\end{tabular}}
\centering
\caption{Performance of pre-trained models from different clarity levels (normal, dark and clear) after they are fine-tuned with different amount of observed data from Lake Mendota. }
\begin{tabular}{l|ccccc}
\hline
\textbf{Method} & \textbf{0\%} & \textbf{0.2\%} & \textbf{2\%}  & \textbf{20\%} & \textbf{100\%} \\ \hline 
RNN$^{EC}$  & - &4.107($\pm$0.181) & 2.149($\pm$0.163) & 1.489($\pm$0.115) &   1.471($\pm$0.077) \\ 
RNN$^{EC,p}$& 2.455($\pm$0.169)& 2.056($\pm$0.180)& 1.590($\pm$0.162)   &   1.402($\pm$0.106)    &1.380($\pm$0.078)\\ 
RNN$^{EC,p\text{(normal)}}$& 2.469($\pm$0.168) & 2.056($\pm$0.184) & 1.595($\pm$0.097) & 1.452($\pm$0.113) & 1.374($\pm$0.074) \\ 
RNN$^{EC,p\text{(dark)}}$& 2.776($\pm$0.124) & 2.067($\pm$0.155) &1.601($\pm$0.078) & 1.393($\pm$0.091)  & 1.380($\pm$0.068) \\ 
RNN$^{EC,p\text{(clear)}}$& 2.518($\pm$0.135) & 2.050($\pm$0.120) & 1.648($\pm$0.128)  & 1.399($\pm$0.088) &1.371($\pm$0.076) \\ \hline 
\end{tabular}
\label{clarity}
\end{table}

\begin{figure} [!h]
\centering
\subfigure[]{ \label{m1}
\includegraphics[width=0.4\columnwidth]{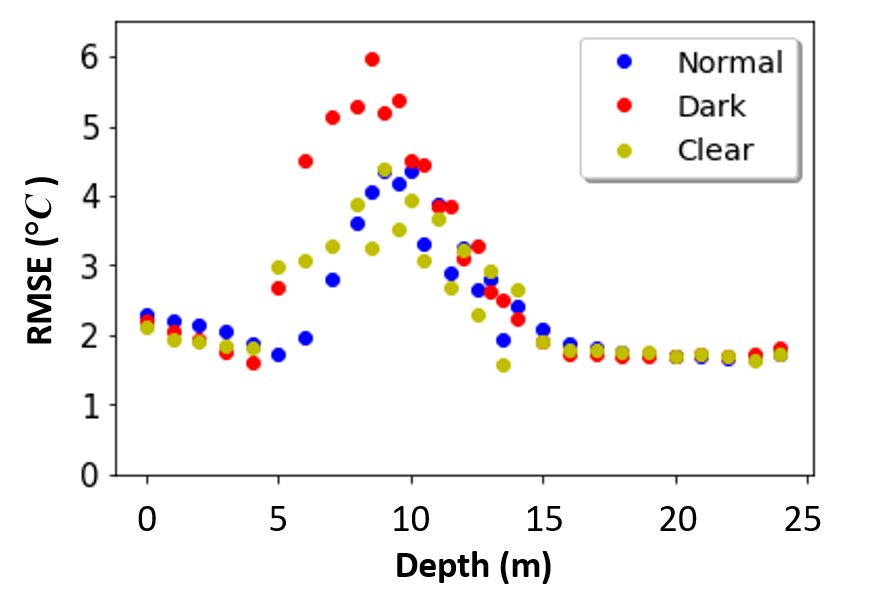}
}
\subfigure[]{ \label{m2}
\includegraphics[width=0.4\columnwidth]{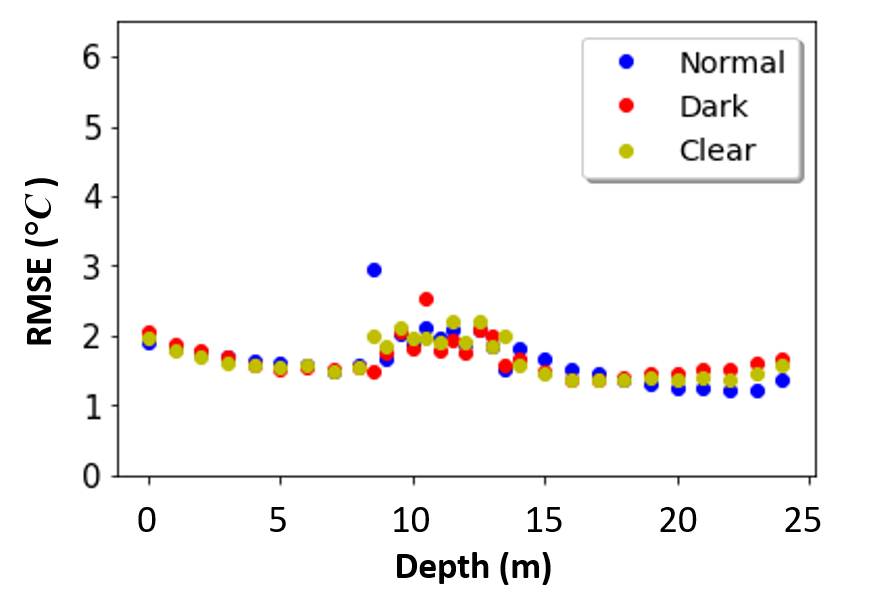}
}
\caption{Error over different depths for pre-trained models under different clarity conditions and fine-tuned with (a) no observed data and (b) 2\% observed data in Lake Mendota.}
\label{err_clarity}
\end{figure}

The water clarity mainly determines how rapidly sunlight is attenuated with respect to water depth. This parameter therefore affects the gradient of the temperature transition and the warming rates of deeper waters. To further analyze this impact, we measure the error across different depths for  models pre-trained under different clarity levels, as shown in Fig.~\ref{err_clarity} (a). It can be seen that the model pre-trained under "dark" clarity has much higher error at depths 6m-12m, where the temperature changes most rapidly. This confirms that a different clarity level can negatively impact water temperature modeling across depths.  However, when we fine-tune the models with a small amount of true observed data, e.g., 2\% data, the model can quickly recover to reasonable performance, as shown in Fig.~\ref{err_clarity} (b). Here it can be seen that the model pre-trained  under "dark" clarity achieves similar performance with models pre-trained under other clarity levels across all the depths. 

\noindent{\underline{Climate conditions:}} 

Next, we generate GLM simulations for a synthetic lake with input drivers from Florida (which are very different from the typically much colder conditions in Wisconsin) and then pre-train the  RNN$^{EC}$ using the simulated data from GLM based on these input drivers. We show the performance of pre-trained models (RNN$^{EC,p\text{(FL)}}$) in Table~\ref{general}. Note that  RNN$^{EC,p\text{(FL)}}$ trained using these input drivers and simulated data in Florida have very poor performance when directly applied to Lake Mendota (9.106 for RNN$^{EC,p\text{(FL)}}$). This is not surprising because there is a huge temperature difference between Wisconsin (where Lake Mendota is located) and Florida. It is more interesting to see that even with just 2\% observations, the learned model becomes much better after fine-tuning. 

\begin{table}[!h]
\small
\newcommand{\tabincell}[2]{\begin{tabular}{@{}#1@{}}#2\end{tabular}}
\centering
\caption{Performance of pre-trained models from Florida (RNN$^{EC,p\text{(FL)}}$) after they are fine-tuned with different amount of observed data from Lake Mendota. We include the performance of RNN$^{EC}$ and RNN$^{EC,p}$ for better comparison. }
\begin{tabular}{l|ccccc}
\hline
\textbf{Method} & \textbf{0\%} & \textbf{0.2\%} & \textbf{2\%}  & \textbf{20\%} & \textbf{100\%} \\ \hline 
RNN$^{EC}$  & - &4.107($\pm$0.181) & 2.149($\pm$0.163) & 1.489($\pm$0.115) &   1.471($\pm$0.077) \\ 
RNN$^{EC,p}$& 2.455($\pm$0.169)& 2.056($\pm$0.180)& 1.590($\pm$0.162)   &   1.402($\pm$0.106)    &1.380($\pm$0.078)\\ 
RNN$^{EC,p\text{(FL)}}$& 9.106($\pm$0.172) & 2.601($\pm$0.177)& 1.759($\pm$0.147)&  1.470($\pm$0.091)  &    1.394($\pm$0.071) \\ \hline  
\end{tabular}
\label{general}
\end{table}

\section{Related Work}
Various components proposed in this work, including generalizing the loss function to include physical constraints, addressing the imperfection of existing physical models, and training ML models using the outputs from physical models, have been studied in different contexts.

As discussed in~\cite{karpatne2017theory}, the idea of including an additional term in the loss function to prefer solutions that are consistent with domain specific knowledge is beginning to find extensive use in many applications. In addition to favoring solutions that are physically consistent, this also allows training in absence of labels, since physics-based loss can be computed even in absence of class labels. Some recent applications of this approach to combining physical knowledge in machine learning can be found in computer vision \cite{sturmfels2018domain, shrivastava2012constrain}, natural language processing \cite{kotzias2015group}, object tracking \cite{stewart2017label}, pose estimation \cite{ren2018learning}, and image restoration \cite{pan2018physicsgan, li2019heavy}. To the best of our knowledge, our work demonstrates for the first time that an ML framework can be adapted to incorporate energy conservation constraint, which is a universal law that applies to many dynamical systems. 

In the context of directly addressing the imperfection of physical models, which is the focus of this paper, the most common approach is residual modeling, where an ML model is learned to predict the errors made by a physics-based model. 
This ML model can be learned using standard supervised learning techniques as long as some observations are available (that can be used to compute the errors made by the physics model).  Once learnt, this ML model is used to make corrections to the output of the physics model.  Most of the work on residual modeling going back several decades (perhaps even earlier) has used plain regression models~\cite{forssell1997combining,xu2015data}, although some recents works~\cite{wan2018data} have used LSTM. A key limitation of such approaches is that they cannot enforce physics based constraints because they try to model the error made by a physics model as opposed to predicting some physical quantity.   Recently, Karpatne et al. introduced a novel hybrid ML and physics model  in which the output of a physics model is fed into an ML model along with inputs that are used to {drive} the physics model~\cite{karpatne2017physics}.  This hybrid model learns to use the output of the physics model as the final output for the input drivers for which physics model is doing well, and make corrections where it makes mistakes. Since the output of this hybrid model is a physical quantity, physics based constraints can now be enforced, allowing for label free learning.   However, such approaches cannot be used to initialize the ML model using just synthetic outputs from the physics model (which are technically free to to obtain)  since they require observations  to be available during training.

Machine learning models are increasingly being used to emulate physics based models since an ML model is typically much faster to execute than a physics based model once it has been trained~\cite{butler2018machine,ojika2017accelerating,mcgregor2017flarenet}.  Since these ML models are trained using synthetic outputs generated by physics based models, the availability of training data is not a limitation, which makes it possible to train even highly complex ML models.  However these emulators (if well trained) can,  in general, be expected to do only as well as the physics models used for generating the training data.  In particular, they cannot correct the errors made by physics-based models due to missing physics or incorrect parameterization. However, the PGRNN approach presented in this paper can be used to develop emulators that are physically consistent and thus likely to more robust and generalizable to out of sample  scenarios.





Another technique to fuse physical models with machine learning is to replace part of the physical model that is costly or inaccurate with a data-driven solution \cite{yao2018chem, tartakovsky2018learning}. In \cite{hamilton2017hybrid}, a subset of the mechanistic model's equations are replaced with data-driven nonparametric methods to improve prediction beyond the baseline process model. As another example from the domain of fluid dynamics,  \cite{raissi2018hidden} uses neural networks to approximate latent quantities of interest like velocity and pressure in Navier Stokes equations. This creates a much more generalizable fluid dynamics framework that doesn't depend as heavily on careful specification of the geometry, as well as initial and boundary conditions. Such approaches are orthogonal to the ones being discussed in our work, as these ML models being used as surrogates can be made "physics-guided" using the framework described in this paper.

There also exists extensive literature on the data-driven discovery of governing equations or mathematical forms that underly complex dynamical systems~\cite{crutchfield1987equations, bongard2007automated, majda2012, sugihara2012detect, brunton2016discovering, raissi2017inferring, raissi2018multi}, or even how to discover the underlying physical laws expressed by partial differential equations from data \cite{raissi2018deep}. 
For example, Rudy et al.~\cite{rudy2017data} present a sparse regression method for identifying governing PDEs from a large library of  potential candidate fictions and spatial-temporal measurements from a  model dynamical system.  Such approaches can be very valuable for analyzing and understanding complex systems for which analytical descriptions are not available (e.g., epidemiology, finance, neuroscience). In contrast,  the  focus of our work is on systems where the dominant governing equations and laws  are already known,  but physics-based models contain inherent  biases, as they are necessarily approximations of reality.

\section{Conclusions}
The PGRNN approach presented in this paper is unique in that it provides a powerful framework for modeling spatial and temporal physical processes while incorporating energy conservation. We also studied the ability of pre-training these models using simulated data to deal with the scarcity of observed data. Using the simulated data from a {generic} physics-based model, PGRNN obtains high prediction performance with fewer observation data used for refinement compared with a parameterized physics-based model calibrated using a large number of observations. Thus, PGRNN can leverage the strengths of physics-based models while filling in knowledge gaps by employing state-of-the-art predictive frameworks that learn from data. 

{The PGRNN framework presented in this paper incorporates energy conservation by adding additional states whose values are computed from physical equations. This allows the use of a rich set of constraints beyond those that can be enforced by just considering the output of the model. In particular, it can be used to model other important physical laws in dynamical systems, such as the law of mass conservation and thus can be used in a wide variety of domains such as hydrology and CFD~\cite{willard2020integrating}. The PGRNN model for GLM made use of seven meteorological/ecological variables, which are used extensively in many environmental modeling applications.   However, key elements of the proposed approach  (ability to incorporate physical laws in the loss function, pre-training using simulation data) are applicable  to other modeling applications which may involve a much larger number of variables. In such cases, the ability of the PGRNN framework to create a highly accurate model using only a small number of observations becomes even more valuable. 
} 

The PGRNN framework can also be viewed as a transfer learning method that transfers the knowledge from physical processes to ML models.  Future research needs to determine the types of dynamical systems models for which such an approach will be effective.  It is entirely possible that new architectural enhancements will need to be made to the traditional LSTM framework to incorporate different types of physical laws and to model underlying physical processes that may be interacting at different spatial and temporal scales. Hence, the proposed framework can be applied to a variety of scientific problems such as nutrient exchange in lake systems and analysis of crop field production, as well as engineering problems such as auto-vehicle refueling design. Therefore, we anticipate this work as an important stepping-stone towards applications of machine learning to problems traditionally solved by physics-based models.

\begin{acks}
This work was supported by NSF award 1934721 and {Department of the Interior Midwest Climate Adaptation Science Center}. We thank North Temperate Lakes Long-Term Ecological Research (NSF DEB-1440297) for temperature and lake metadata. Access to computing facilities was provided by Minnesota
Supercomputing Institute. {Reference to trade names does not imply endorsement by the U. S. government.}


\end{acks}

\bibliographystyle{ACM-Reference-Format}
\bibliography{sample-bibliography}

\appendix

\section{Energy conservation }

\begin{figure} [!h]
\centering
\label{fig:b}{}
\includegraphics[width=0.3\columnwidth]{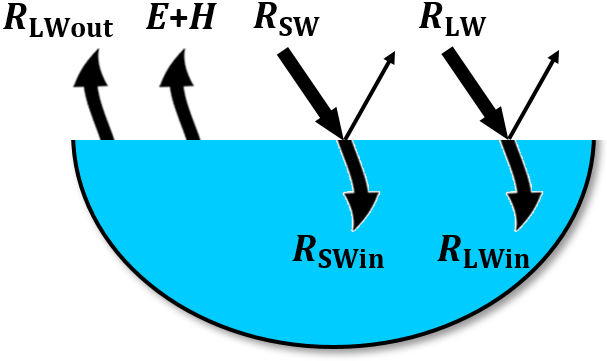}
\caption{The heat energy fluxes that are modeled in PGRNN. For short-wave radiation ($R_{\text{SW}}$) and long-wave radiation ($R_{\text{LW}}$), a portion of the energy is reflected by the lake surface.}
\label{energy}
\end{figure}

In Fig.~\ref{energy}, we show the major incoming and outgoing heat fluxes that impact the lake energy. The incoming heat fluxes include terrestrial long-wave radiation and incoming short-wave radiation. The lake loses heat mainly through the outward fluxes of back radiation ($R_{\text{LWout}}$), sensible heat fluxes ($H$), and latent evaporative heat fluxes ($E$)\footnote{Here the latent heat fluxes are related to changes in phase between liquids, gases, and solids while the sensible heat fluxes are related to changes in temperature  with no change in phase~\cite{bruce2018multi}.}.

We now expand  Eq.~\ref{law_detail} with more detailed energy fluxes. 
The consistency between lake energy $U_t$ and detailed energy fluxes can be expressed as:
\begin{equation}
\small
    \Delta U_t = R_{\text{SW}} (1-\alpha_{\text{SW}})+R_{\text{LWin}} (1-\alpha_{\text{LW}})-R_{\text{LWout}}-E-H,
\label{law_detail}
\end{equation}
where $\Delta U_t=U_{t+1}-U_t$, $\alpha_{\text{SW}}$ is the short-wave albedo (the fraction of short-wave energy reflected by the lake surface) and $\alpha_{\text{LW}}$ is the long-wave albedo. In our implementation, we set $\alpha_{\text{SW}}$ to 0.07 and $\alpha_{\text{LW}}$ to 0.03 which are generally accepted values for lakes from previous scientific studies~\cite{hipsey2019general}.  
All energy components are in Wm$^{-2}$. By comparing this with Eq.~\ref{law_detail}, we can see that $\mathcal{F}_{in}=R_{\text{SW}} (1-\alpha_{\text{SW}})+R_{\text{LWin}} (1-\alpha_{\text{LW}})$ and $\mathcal{F}_{out} = R_{\text{LWout}}+E+H$. In this work, we ignore the smaller flux terms such as sediment heat flux and advected energy from surface inflows and groundwater.

\vspace{.1in}
\noindent\textbf{Estimation of Heat Fluxes and Lake Thermal Energy:} 
We now introduce how to estimate energy fluxes in our implementation.

Terrestrial long-wave ($R_{\text{LWin}}$) radiation is emitted from the atmosphere, and depends on prevailing local conditions like air temperature and cloud cover. Incoming short-wave radiation ($R_{\text{SW}}$) is affected mainly by latitude (solar angle), time of year, and cloud cover. Both factors are included in the input drivers $X$.

As for the outgoing energy fluxes, we estimate $E$, $H$, and $R_{\text{LWout}}$ separately using the input drivers and modeled surface temperature.  

The sensible heat flux and latent evaporative heat flux can be computed based on the previous study~\cite{hipsey2019general}: 
\begin{equation}
\small
\begin{aligned}
    E &= -\rho_a C_E \nu \kappa_{10} \frac{\omega}{p} (e_s-e_a),\\
    H &= -\rho_a c_a C_H \kappa_{10} (T_s-T_a),
\end{aligned}
\label{eh}
\end{equation}
where $C_H$ is the bulk aerodynamic coefficients for sensible heat transfer, and $C_E$ the bulk aerodynamic coefficients for latent heat transfer. Both coefficients are estimated from Hicks' collection of ocean and lake data~\cite{hicks1972some}. The coefficient $\omega$ is the ratio of the molecular mass of water to the molecular mass of dry air (=0.622), $\nu$ the latent heat of vaporization (=2.453$\times$10$^6$), and $c_a$ the specific heat capacity of air (=1005). 
The variable $T_a$ is the air temperature, and $\kappa_{10}$ the wind speed (m/s) above the lake referenced to 10m height. Both these variables are included or can be derived from input drivers. $T_s$ is the surface water temperature in {Kelvin} obtained through the feed-forward process. The air density $\rho_a$ is computed as $\rho_a = 0.348 (1+r) /(1+1.61 r) p/T_a$, where $p$ is air pressure (hPa) and $r$ is the water vapour mixing ratio (both derived from input drivers). The vapour pressure ($e_s$ and $e_a$) is calculated by the linear formula from Tabata~\cite{tabata1973simple}:
\begin{equation}
\small
    \begin{aligned}
    e_s &= 10^{(9.28603523\frac{2322.37885}{T_s+273.15})},\\
    e_a &= (S_{\text{RH}}RH/100)e_s,
    \end{aligned}
    \label{esa}
\end{equation}
where $S_{\text{RH}}$ is the relative humidity scaling factor (=1, obtained through calibrating the GLM model) and $RH$ is the relative humidity (included in input drivers).

The back radiation  $R_{\text{LWout}}$ is estimated as: 
\begin{equation}
\small
    \begin{aligned}
    &R_{\text{LWout}} = \epsilon_s \delta T_s^4,\\
    \end{aligned}
    \label{rout}
\end{equation}
where $\epsilon_s$ is the emissivity of the water surface (=0.97), and $\delta$ is the Stefan-Boltzmann constant (=5.6697e-8 Wm$^{-2}$K$^{-4}$). 

\section{Distribution of training data}

\begin{figure} [!h]
\centering
\label{fig:b}{}
\includegraphics[width=0.6\columnwidth]{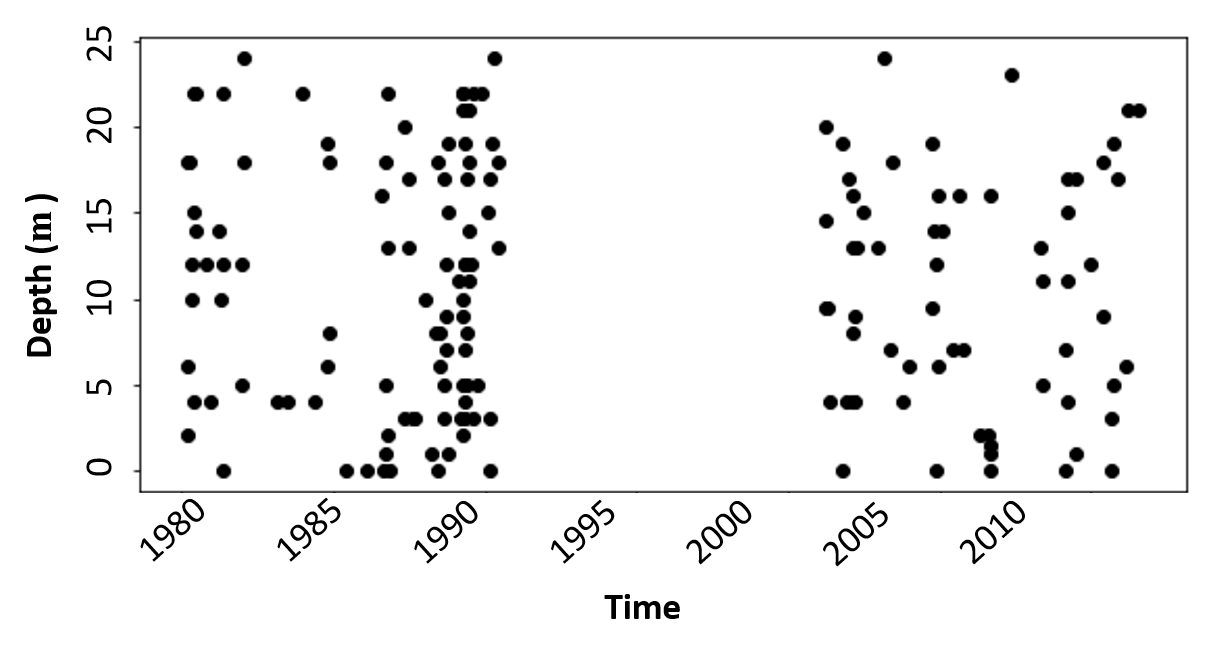}
\caption{Distribution of 2\% observed data used for training.}
\label{dist_2p}
\end{figure}

In Fig.~\ref{dist_2p}, we show the distribution of randomly selected 2\% observed data across different depths and different dates. 

\end{document}